\documentclass[journal,twoside,web]{ieeecolor}
\usepackage{generic}
\usepackage{cite}
\usepackage{amsmath,amssymb,amsfonts}
\usepackage{graphicx}
\usepackage{algorithm,algorithmic}
\usepackage{hyperref}
\hypersetup{hidelinks}
\usepackage{textcomp}
\usepackage{booktabs}
\usepackage{tabularx}
\usepackage{array}
\usepackage{placeins}
\providecommand{\refname}{References}
\usepackage[caption=false,font=footnotesize]{subfig}
\def\BibTeX{{\rm B\kern-.05em{\sc i\kern-.025em b}\kern-.08em
    T\kern-.1667em\lower.7ex\hbox{E}\kern-.125emX}}
\begin{document}
\title{Machine Learning-Based Real-Time Detection of Compensatory Trunk Movements Using Trunk-Wrist Inertial Measurement Units}
\author{
Jannis~Gabler$^{*}$, 
Cl\'ement~Lhoste$^{*}$,
Max~Quast,
Laura~Mayrhuber,
Andrea~Ronco,
Olivier~Lambercy$^{\dagger}$,
Paulius~Viskaitis$^{\dagger}$,~and~Dane~Donegan$^{\dagger}$
\thanks{This work was supported by Bridge Proof of Concept 40B1-0 214621, Pioneer fellowship PIO-03 22-2, Gebert Rüf Stiftung GRS-032/23, Innosuisse 113.845 IP-LS programs and ZNZ PhD Grant. Additionally, part of this research is supported by the National Research Foundation Singapore (NRF) under its Campus for Research Excellence and Technological Enterprise (CREATE) program.}%
\thanks{*Jannis Gabler and Cl\'ement Lhoste contributed equally to this work.}%
\thanks{$^{\dagger}$Olivier Lambercy, Paulius Viskaitis, and Dane Donegan contributed equally to the conception and scientific supervision of this study and share last authorship.}%
\thanks{J. Gabler is with the Rehabilitation Engineering Laboratory, ETH Zurich, Switzerland, and also with TU Munich, Germany.}%
\thanks{C. Lhoste is with the Rehabilitation Engineering Laboratory, ETH Zurich, Switzerland, and the Neuroscience Center Zurich (ZNZ), Switzerland.}%
\thanks{M. Quast, L. Mayrhuber, A. Ronco, and P. Viskaitis are with the Rehabilitation Engineering Laboratory, ETH Zurich, Switzerland.}%
\thanks{O. Lambercy is with ETH Zurich, Switzerland, the Neuroscience Center Zurich (ZNZ), Switzerland, and the Singapore-ETH Centre (CREATE), Singapore.}%
\thanks{D. Donegan is with the Rehabilitation Engineering Laboratory, ETH Zurich, Switzerland (corresponding author: dane.donegan@hest.ethz.ch).}%
}

\maketitle

\begin{abstract}
Compensatory trunk movements (CTMs) are commonly observed after stroke and can lead to maladaptive movement patterns, limiting targeted training of affected structures. Objective, continuous detection of CTMs during therapy and activities of daily living remains challenging due to the typically complex measurements setups required, as well as limited applicability for real-time use. This study investigates whether a two-inertial measurement unit configuration enables reliable, real-time CTM detection using machine learning. Data were collected from ten able-bodied participants (ETH Zurich Ethics Commission, EK 2022-N-195) performing activities of daily living under simulated impairment conditions (elbow brace restricting flexion-extension, resistance band inducing flexor-synergy-like patterns), with synchronized optical motion capture (OMC) and manually annotated video recordings serving as reference. A systematic location-reduction analysis using OMC identified wrist and trunk kinematics as a minimal yet sufficient set of anatomical sensing locations. Using an extreme gradient boosting classifier (XGBoost) evaluated with leave-one-subject-out cross-validation, our two-IMU model achieved strong discriminative performance (macro-F1 = 0.80 $\pm$ 0.07, MCC = 0.73 $\pm$ 0.08; ROC-AUC $>$ 0.93), with performance comparable to an OMC-based model and prediction timing suitable for real-time applications. Explainability analysis revealed dominant contributions from trunk dynamics and wrist–trunk interaction features. In preliminary evaluation using recordings from four participants with neurological conditions (BASEC ID: 2025-01124 and 2024-01425), the model retained good discriminative capability (ROC-AUC $\approx$ 0.78), but showed reduced and variable threshold-dependent performance, highlighting challenges in clinical generalization. These results support sparse wearable sensing as a viable pathway toward scalable, real-time monitoring of CTMs during therapy and daily living.
\end{abstract}


\section{Introduction}
\label{sec:introduction}
\IEEEPARstart{S}{troke} is a leading cause of long-term upper-limb disability, with over 50\% of survivors experiencing persistent sensorimotor deficits six months post injury  \cite{oflaherty_2024}. In response, physical and occupational therapy aims to restore functional abilities for activities of daily living (ADL), yet patients often adopt compensatory movements (CMs) to offset the limited range of motion in the affected limb \cite{levin_2009, jones_2017}. One prevalent example are compensatory trunk movements (CTMs), in which trunk flexion, side lean, or rotation are used to compensate for deficits in elbow extension or shoulder flexion, often arising from abnormal flexor synergies \cite{bokkyu_2024, murphy_2015, brami_2003}. Although CTMs can facilitate functional ability during task performance in the short term, they reinforce potentially maladaptive movement patterns and limit targeted training of affected structures, thereby contributing to persistent motor dysfunction and secondary musculoskeletal pain \cite{jones_2017}. Consequently, accurate monitoring and assessment of CTMs are essential to guide rehabilitation more effectively.

In this context, standardized clinical assessments such as the Fugl-Meyer Upper Extremity Assessment (FMA-UE) \cite{see_2013} and the Wolf Motor Function Test (WMF) \cite{wolf_2001} can indirectly inform on CMs, while the Reaching Performance Scale (RPS) directly scores CTMs \cite{levin_2004}. Although widely used, these assessments rely on trained raters, can suffer from inter-rater variability \cite{duff_2014}, and are typically administered infrequently in supervised clinical environments. As a result, there is a lack of continuous and objective assessments capable of capturing CTMs as they emerge in naturalistic, real-world environments. This gap has motivated the development of technology-based approaches that could automate CTM detection and provide scalable alternatives to clinic-based evaluations, primarily through camera-based systems and wearable sensors \cite{wang_2021}. Multiple studies have demonstrated the feasibility of CTM detection using low-cost, camera-based skeletal tracking combined with threshold-based approaches \cite{avarell_2022,coias_2022} or machine-learning (ML) algorithms \cite{zhi_2017, nordin_2016, lin_2023, unger_2025}. Marker-less tracking and robustness to sensor drift represent key advantages of these systems. However, susceptibility to occlusions and the need for precise positioning or extensive calibration in multi-camera configurations limit their usability in unsupervised settings and constrain their deployment environments.

Inertial measurement units (IMUs) offer a portable and low-cost alternative for CTM detection. Existing approaches range from systems employing sensors on multiple body segments to enable detailed joint-angle reconstruction \cite{held_2018, schwarz_2020, schwarz_2021, Wittmann2016} to simple configurations relying on a single trunk-mounted sensor \cite{nguyen_2021}. Despite this diversity, existing methods exhibit several limitations. Multi-sensor approaches increase setup effort and potential wearing discomfort, and require more elaborate calibration procedures for accurate joint-angle reconstruction. In contrast, single-sensor approaches, while showing simplicity in use and setup, typically rely on restricted workspaces or comparisons with the non-affected side, constraining their use in real-life scenarios and limiting their applicability for real-time monitoring~\cite{nguyen_2021}.

To address this gap, we introduce a novel framework for real-time detection of CTMs using a sparse wearable sensing setup and ML. The approach is designed to support robust and unobtrusive monitoring of CTMs during therapy and ADL, providing a foundation for future real-time feedback applications for patients and clinicians across settings. The framework leverages a minimal two-IMU configuration (wrist and trunk) and a window-based classification pipeline to continuously infer CTM occurrence during movement execution. We hypothesized that wrist and trunk kinematics provide complementary information sufficient to discriminate CTMs from physiologically appropriate trunk involvement in unconstrained workspaces, and that this minimal sensing configuration enables robust generalization across users and movement tasks. To evaluate the proposed approach, we recorded synchronized IMU and optical motion capture (OMC) data from ten able-bodied participants performing a wide range of ADL under externally imposed constraints using an elbow brace and a resistance band, designed to approximate post-stroke movement limitations. Ground-truth labels of CTMs were obtained through manual annotation of synchronized video recordings. First, using OMC data, we performed a location-reduction analysis to identify a minimal set of anatomical locations enabling reliable CTM detection. This approach isolated location-dependent effects from sensor non-idealities and was evaluated using nested leave-one-subject-out cross-validation (LOSO-CV). In line with these results, a corresponding two-IMU model was trained and similarly evaluated. Real-time feasibility was assessed by implementing the pipeline in real time and measuring end-to-end prediction latency. Finally, preliminary clinical generalization was investigated by testing the model trained on able-bodied data on recordings from four individuals with neurological conditions performing reaching tasks and conventional upper-limb therapy.

\section{Methods}
\label{sec: methods}

\subsection{Able-bodied data collection}
\label{sec:acquisition}
%\textbf{Participants}

%\noindent
Data were collected from ten able-bodied adult participants (mean age: 25.3 ± 2.1 years; 20\% female), all of whom were included in the analyses. Participants
were exclusively recruited from the student population at ETH Zurich. The study was approved by
ETH Zurich Ethics Commission (EK 2022-N-195) and informed consent was obtained from all
participants.

%\noindent
%\textbf{Experimental Setup}

\begin{figure}[H]
    \centering
    \includegraphics[width=1\linewidth]{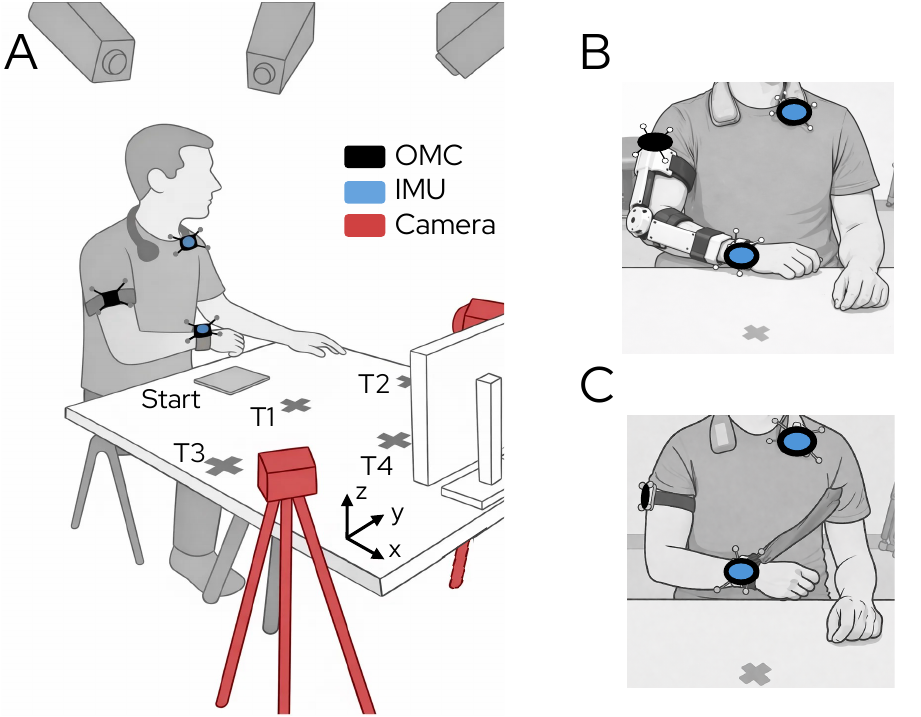}
    \caption{Experimental setup for data acquisition and tasks conditions. (A) Optical motion capture (OMC) markers (black) were attached to the wrist, upper arm, and trunk, while inertial measurement units (IMU, blue) were co-mounted at the wrist and trunk. RGB cameras (red) provided synchronized video recordings from two viewpoints. Participants performed standardized activity of daily living at pre-defined target locations (``Start'', ``T1-T4'') with task instructions displayed on a screen positioned in front of the participant. The axes of the anatomically aligned coordinate frame are indicated (x, y, z). (B) Schematic of the  brace condition used to restrict elbow flexion and extension. (C) Schematic of the band condition designed to induce flexor-synergy like behavior.}
    \label{fig:exp_setup}
\end{figure}

%\noindent
A multimodal data acquisition setup was implemented to enable synchronized kinematic analysis and visual inspection by integrating marker-based OMC, IMUs, and video recordings  (Fig. \ref{fig:exp_setup}A). OMC data was collected using a 14-camera system (OptiTrack, NaturalPoint, Corvallis, USA) operating at 120 Hz and served as the reference standard for movement kinematics. Reflective cluster markers were placed on the wrist, upper arm, and a necklace positioned near the sternum to capture motion of the individual segments. IMU sensors (LSM6DSV16BX, STMicroelectronics, Geneva, Switzerland) were co-located with the wrist and necklace markers, corresponding to the intended final two-IMU configuration, also sampled at 120 Hz. In addition, two webcams positioned at 45° angles in front of the participant recorded synchronized videos at a target frame rate of 15 frames per second, to support manual annotation and visual interpretation of sensor signals.

%%\noindent
%%\textbf{Experimental Protocol}

%%\noindent
To investigate CTM detection in able-bodied participants, tasks were performed under three movement conditions approximating a range of motor behaviors from unconstrained to stroke-like patterns: (A) unrestricted arm movement allowing natural motion; (B) constrained movement using an elbow brace limiting flexion and extension, reflecting reduced elbow mobility commonly observed after stroke (Fig. \ref{fig:exp_setup}B)~\cite{beer_1999, mccrea_2002}; and (C) movement against a diagonally oriented resistance band extending from the contra-lateral shoulder to the wrist, inducing flexor-synergy-like behavior frequently reported in stroke populations (Fig. \ref{fig:exp_setup}C)~\cite{dewald_1995, ellis_2017}. Building on these movement conditions, the protocol emphasized movement diversity over repeated execution of a limited task set, to better reflect the natural variability of ADL-related upper-limb movements. Tasks were adapted from established motor assessment procedures \cite{wolf_2001, levin_2004,johansson_2015, lyle_1982, jebsen_1969} and grouped into six categories: (1) Planar Reaching \& Transport, (2) Drinking \& Pouring, (3) Elevated Reaching \& Transport, (4) Planar Sliding, (5) Fine Object Manipulation, and (6) Continuous Movements. Within each category, a set of distinct movement tasks was defined (38 in total across all categories), with each task performed once under each of the three movement conditions (see Appendix~Tab. \ref{tab:tasks} for more detail). Movements were performed using the right arm to reduce data collection requirements.%, while sagittal planar mirroring of the recorded signals allowed access to left-arm–like data.

Prior to the experiment, participants were seated upright on a non-rotating chair. Table height and distance were individually adjusted to support a natural posture. Individual reaching distance (defined as the maximal arm extension achievable without trunk involvement) was measured and used to define four personalized target locations (Fig. \ref{fig:exp_setup}A; further details in Appendix \ref{sec:appendix_workspace}). Three of these targets locations (T2-T4) were selected to naturally elicit trunk involvement even under the unrestricted condition, ensuring that trunk motion was also present during physiologically appropriate, non-compensatory movements. In contrast, the “Start” and “T1” targets were designed to capture movements that did not require trunk involvement in the unconstrained condition.

At the beginning of each recording, participants performed a standardized calibration pose consisting of an upright trunk posture, approximately 90° elbow flexion and the forearm aligned in the left lateral direction (Fig.~\ref{fig:exp_setup}). This pose was selected due to its comfort and feasibility for individuals with typical post-stroke movement limitations and was later used to establish an anatomically aligned coordinate system for both IMU sensors. Following calibration, participants performed movement tasks based on standardized visual and verbal instructions displayed on a screen (Fig.~\ref{fig:exp_setup}A), with each task starting and ending in the calibration pose. The order of tasks was randomized for each participant to minimize anticipatory behavior and distribute potential sensor drift between task categories. This procedure was repeated for all three movement conditions, which were also randomized in advance. This resulted in uninterrupted multi-modal recordings (13–18 min per condition), reflecting real-world measurement conditions.

\subsection{Clinical dataset}

To complement the able-bodied dataset and assess preliminary performance of the proposed approach with persons with neurological conditions, a clinical dataset was aggregated from two independent studies conducted in heterogeneous settings. Four patients provided written informed consent in accordance with the Declaration of Helsinki.

First, three male chronic stroke patients (P18: 44 years, left-affected, FMA = 18; P31: 65 years, left-affected, FMA = 26; P31: 59 years, right-affected, FMA = 52) were recruited from an ongoing study (BASEC ID: 2025-01124). Similar to Fig. \ref{fig:exp_setup}, synchronized OMC, IMU and video data were collected. IMUs were placed on both wrists, as well as on the trunk. In that study, the patients performed planar reaching and transport tasks (see Appendix~Tab. \ref{tab:tasks}) sequentially with the most affected and less affected arms. During this session, reaching movements were executed toward three target locations (left, right, and front), with five repetitions per location, followed by an additional frontal reaching task that was also repeated five times per arm (around 7min).

In addition, movement data from a patient with spinal cord injury during upper-limb physical therapy (P03: male, 67 years, ARAT: 27), was included and re-analyzed for the purpose of the present study \cite{Lhoste2025}. The corresponding study protocol was approved by the local cantonal medical and ethics committee (BASEC ID: 2024-01425) and registered on ClinicalTrials.gov (NCT06623721). IMUs were placed on the trained arm (right side) and on the trunk, and synchronized video recordings were acquired. Data were collected during the performance of a pegboard task and a cup-stacking task, representing activities distinct from those included previously to assess generalizability. A continuous 10-min segment was selected for analysis.

\subsection{Data annotation}
\label{sec:annotation}
To support supervised model training and performance evaluation using IMU and OMC data, a ground-truth reference was established through manual annotation of synchronized video recordings. Based on the expected biomechanical effects of the experimental constraints (elbow brace and resistance band), which primarily limited elbow flexion and extension, trunk flexion was anticipated to be the dominant CTM. Occasional trunk rotation or lateral flexion (typically observed when shoulder abduction or rotation is limited) was observed, but grouped under a single CTM class to avoid severe class imbalance. Three classes were defined: (1) Calibration (Calib), (2) Movement: No Trunk Compensation (Mov: No TC), and (3) Movement: Trunk Compensation (Mov: TC). Importantly, labels were assigned based on the movement context rather than instantaneous motion; brief pauses during task execution (e.g., when reaching a target or placing an object) were still considered part of the movement phase and annotated as Mov: No TC or Mov: TC, rather than Calib.

\begin{figure}[!t]
    \centering

    \subfloat[]{
        \includegraphics[width=0.42\linewidth]{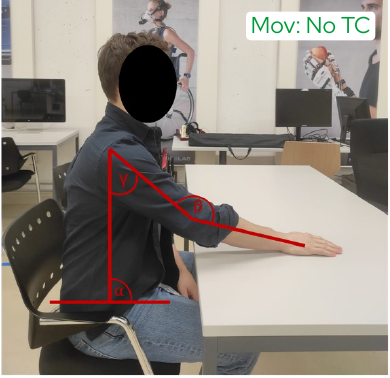}
        \label{fig:labeling_a}
    }
    \hfill
    \subfloat[]{
        \includegraphics[width=0.42\linewidth]{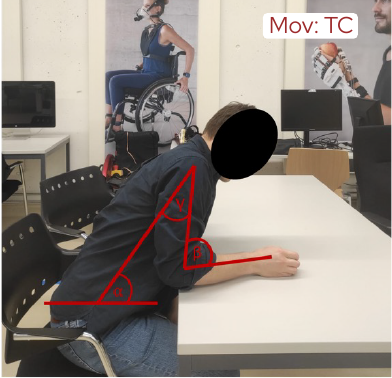}
        \label{fig:labeling_b}
    }

    \vspace{1mm}

    \subfloat[]{
        \includegraphics[width=0.42\linewidth]{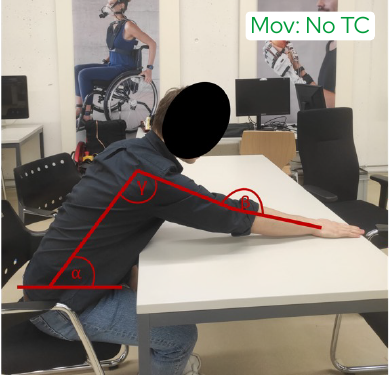}
        \label{fig:labeling_c}
    }
    \hfill
    \subfloat[]{
        \includegraphics[width=0.42\linewidth]{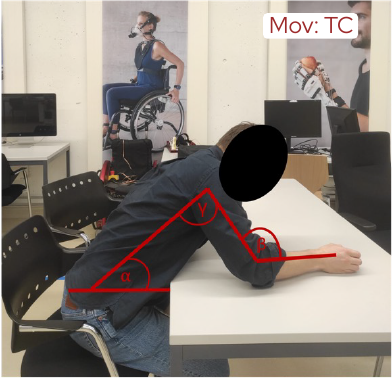}
        \label{fig:labeling_d}
    }

    \caption{Example scenarios illustrating arm--trunk coordination strategies during a forward-reaching task and the rationale for labeling compensatory trunk movements. 
    (a) The degree of elbow extension ($\beta$) and shoulder flexion ($\gamma$) is sufficient to reach the target. Trunk flexion is therefore not required to extend the arm's reach.
    (b) The degree of elbow extension ($\beta$) and shoulder flexion ($\gamma$) is insufficient to reach the target. In response, the trunk is flexed ($\alpha \downarrow$) to extend the arm's forward reach.
    (c) The degree of elbow extension ($\beta$) and shoulder flexion ($\gamma$) is maximized. The trunk is flexed to extend the arm's forward reach toward the target. Since this trunk flexion is not compensating for limited elbow extension or shoulder flexion, it is not considered compensatory.
    (d) The degree of elbow extension ($\beta$) and shoulder abduction ($\gamma$) stays significantly below the required range to reach the target. Therefore, the trunk is excessively flexed ($\alpha \downarrow$) as a compensatory strategy to extend the arm's forward reach.
    Red annotations indicate trunk flexion angle ($\alpha$), elbow extension ($\beta$), and shoulder flexion/abduction ($\gamma$).}
    \label{fig:labeling}
\end{figure}

Labeling guidelines were established to formalize these classes in a task-agnostic and workspace-independent manner (see Appendix~\ref{sec:appendix_labeling}). Although informed by existing literature \cite{barth_2020, sauerzopf_2024}, the guidelines were refined to emphasize the relationship between trunk and wrist movement, following principles from the RPS \cite{levin_2004}. The core idea is that trunk motion is not inherently compensatory: large trunk excursions can be non-compensatory when required by the task or workspace and accompanied by appropriate arm extension, whereas small trunk movements can be compensatory when they substitute for insufficient elbow extension and are used to achieve the required range of motion (Fig.~\ref{fig:labeling}). This definition is particularly relevant for real-world, out-of-laboratory settings, where tasks are unconstrained and can naturally elicit trunk involvement even during physiologically appropriate movements. Accordingly, this labeling strategy aligns with the previously described target positions, which necessarily involve trunk contribution even in the unimpaired condition (see Sec.~\ref{sec:acquisition}). Since we aimed to establish a CTM framework producing continuous predictions during movement execution rather than a single label per movement sequence, the guidelines also specified rules for class onsets and offsets (e.g., when in time a CTM starts and stops). All criteria were reviewed and refined by experienced physiotherapists.

For annotation, synchronized dual-view videos were stacked to reduce occlusions and ensure consistent visibility of trunk and arm kinematics. The combined videos were uploaded to Labelbox (Labelbox, Inc., San Francisco, USA), where they were annotated frame-by-frame following the finalized guidelines, with exactly one class assigned to each frame. All able-bodied recordings were labeled by one trained researcher to ensure consistency (JG), while the clinical dataset was annotated by a researcher with a background in physical therapy and experience working with neurological patients (LM). %and a subset was independently annotated to assess inter-rater reliability.
%Abled-bodied dataset was labeled by train researchers, 

\subsection{Data processing and model training}
\label{sec:processing_training}
%\noindent
%\textbf{Pre-Processing:}

%\noindent
For each recording, raw accelerometer and gyroscope signals from the wrist and trunk IMUs were bias-corrected using predefined, axis-specific values to reduce susceptibility to drift. Then, a 6D Versatile Quaternion Filter (VQF) \cite{ladig_2022} was used to estimate sensor orientations in a global world frame and allow gravity to be removed from the accelerometer data. To align this world frame with participants’ anatomical axes, a static anatomical calibration was performed using the known Calib posture at the start of each recording (see Appendix~\ref{sec:appendix_calibration}). The orientation of this anatomically aligned coordinate frame is shown in Fig.~\ref{fig:exp_setup}A (x: anterior, y: left-lateral, z: superior). This procedure enabled consistent interpretation of orientation data relative to the workspace while allowing both native sensor-frame data and anatomically aligned representations of acceleration and angular velocity to be used (see Appendix~\ref{sec: kin_streams}). %Hereafter, we refer to the native sensor frame as the \textit{local frame} and to the anatomically aligned frame obtained through calibration as the \textit{calibrated frame}. 
For patients, the Calib pose was also accepted with arms resting on the table and aligned with the sagittal plane; left-arm recordings were subsequently mirrored with respect to the sagittal plane to obtain right-arm–equivalent data (see Appendix~\ref{sec:appendix_calibration} for details).

The OMC data provided rigid-body marker positions and segment orientations as unit quaternions relative to the laboratory reference frame defined during system setup. This reference frame was configured to match the anatomically aligned IMU frame, enabling direct cross-modal comparison of kinematic variables. %and is hereafter referred to as the \textit{recording frame}. 
The two modalities were temporally aligned by maximizing the cross-correlation between the corresponding IMU and OMC marker orientation signals. From the synchronized OMC and IMU signals, additional kinematic data streams were derived for downstream analysis, comprising motion-related features (acceleration, angular velocity, and linear velocity), pose-related features (absolute segment orientation and relative inter-segment orientation), and spatial features based on OMC marker distances (see Appendix~\ref{sec: kin_streams}). For subsequent ML model training, video-based ground-truth labels were exported from Labelbox and mapped to both IMU and OMC time-series using nearest-neighbor interpolation, resulting in a fully synchronized and labeled multimodal dataset. 

%\noindent
%\textbf{Model Training and Evaluation Pipeline:}

\begin{figure*}[t]
    \centering
    \includegraphics[width=0.9\textwidth]{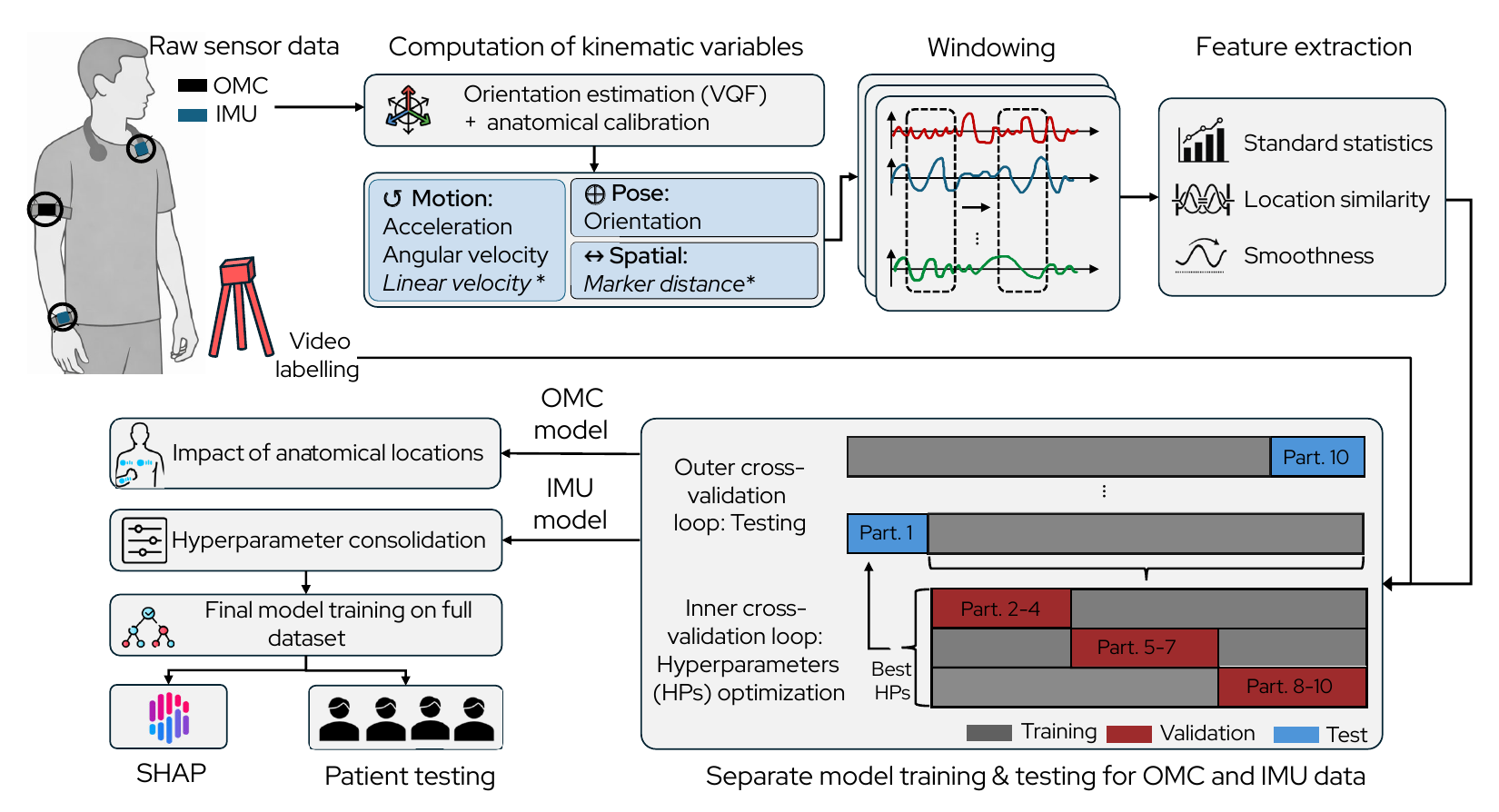}
    \caption{Overview of the signal processing and modeling pipeline. Inertial measurement units (IMU) and optical motion capture (OMC) signals are transformed into multiple kinematic representations (*variables in italic are only present in OMC data), segmented into overlapping windows, and characterized using statistical, location similarity (pairwise comparison of kinematic streams to capture coordination, e.g., wrist–trunk), and smoothness metrics. Model training and evaluation are conducted within a nested leave-one-subject-out cross-validation (LOSO-CV) scheme with balanced sample weighting to mitigate class imbalance. Final evaluation on patient data, together with post-hoc SHAP analysis \cite{shap_2017}, supports preliminary clinical applicability and provides insight into model interpretability.}
    \label{fig:pipeline}
\end{figure*}

%\noindent
Based on the pre-computed kinematic data streams, we designed a pipeline for training and evaluating our CTM-prediction model (Fig.~\ref{fig:pipeline}). Each recording was segmented into sliding windows of 500 ms (60 samples) with a 375 ms overlap (75\% of the window size). For each window, a set of features consisting of statistical, location similarity (pairwise comparison of kinematic streams to capture coordination, e.g., wrist–trunk), and smoothness metrics (see Appendix~\ref{sec: kin_streams}) was applied on the kinematic data streams, forming the input of our model. These features were designed to capture both kinematic amplitude and inter-segment coordination, as well as movement quality, since CTMs are hypothesized to manifest as altered coordination patterns and reduced smoothness compared to physiologically appropriate motion. The label for each window was defined as the last sample label within the window, reflecting a causal prediction setting in which the model infers the most recent movement state from past data with minimal latency. Predictions were obtained by selecting the class with the highest posterior probability. Overall, this windowing strategy enables frequent predictions while retaining sufficient temporal context.

As a classifier, extreme gradient boosting (XGBoost) was selected due to its demonstrated performance and computational efficiency on tabular, hand-crafted feature sets \cite{chen_2016}. To assess subject-independent generalization and limit the risk of overfitting given the large hyperparameter (HP) space of XGBoost, a nested LOSO-CV scheme was implemented (Fig.~\ref{fig:pipeline}). For this, one fold (comprising all data from one participant) served as the test set. Data from the remaining nine participants were used in an inner 3-fold CV for HP optimization. The search space was kept constant across folds, and hyperparameters were optimized using Bayesian optimization over 50 iterations. To account for the inherent class imbalance in our dataset (Calib: 34.51\%, Mov: No TC: 48.84\%, Mov: TC: 16.65\%), we employed balanced sample weights to mitigate bias toward majority classes during training. After identifying the optimal HPs, the model was retrained on data from the nine participants and tested on the held-out participant. This procedure was repeated until each participant had served once as the test fold.

\subsection{Data analysis}
Using the proposed pipeline, a series of experiments was conducted to systematically evaluate model behavior and performance under different experimental settings (Fig.~\ref{fig:pipeline}). Across all experiments, classification performance was primarily quantified using F1-scores and Matthews correlation coefficient (MCC) as a complementary metric that is robust to class imbalance and accounts for all entries of the confusion matrix \cite{Chicco2020}. As secondary metrics, we used one-vs-rest receiver operating characteristic (ROC) curves with their corresponding area under the curves (AUC) values, precision and recall, as well as confusion matrices. For the LOSO-CV evaluation, performance metrics were computed independently for each participant and summarized as mean ± standard deviation across participants, whereas confusion matrices were aggregated across all test participants.

First, to test the hypothesis that wrist- and trunk-derived kinematics provide complementary information sufficient for CTM detection, we progressively reduced the marker set of the OMC-derived model and quantified the resulting change in classification performance. This analysis was performed on drift-free OMC data to isolate the effect of anatomical location from IMU-specific non-idealities such as drift and calibration inaccuracies. To statistically validate performance differences, paired Wilcoxon signed-rank tests were performed on per-subject F1 scores of the Mov: TC class, as this class reflects CTM detection performance and is therefore of primary clinical relevance. A set of planned comparisons was defined a priori. First, a comparison between the three-marker (wrist+trunk+upper arm) and two-marker (wrist+trunk) configurations was performed to assess the contribution of upper-arm kinematics. Subsequently, two additional comparisons were conducted to evaluate the effect of reducing the configuration to a single marker (wrist+trunk vs.\ wrist and wrist+trunk vs.\ trunk). To control for multiple comparisons, the Holm–Bonferroni correction was applied to control the family-wise error rate. Adjusted p-values are reported, and statistical significance was defined as $p_{corr}<$ 0.05 across the manuscript. For completeness, the same statistical analysis was also performed for all classes (Calib, Mov: No TC, Mov: TC) and all pairwise model comparisons.

Second, we trained a model using the proposed two-IMU (wrist+trunk) configuration and evaluated it on able-bodied data, directly comparing it against the baseline OMC-derived two-marker (wrist+trunk) model. Paired differences between OMC and IMU performance for each class F1-scores were assessed using Wilcoxon signed-rank tests, with Holm–Bonferroni correction applied to control the family-wise error rate across multiple comparisons. In addition, we included an analysis of task-dependent performance variations to assess the robustness of the model across different movement contexts and identify potential task-specific performance differences, to reflect the broad range of tasks encountered in daily-life. Following this evaluation, a final IMU model was trained using all available able-bodied data, without a separate test set, for subsequent analyses. Hyperparameters obtained from the nested LOSO-CV were consolidated by aggregating numerical parameters using the median and categorical parameters using the most frequently selected value. This resulting model was subsequently subjected to an explainability analysis using Shapley Additive Explanations (SHAP) \cite{shap_2017} to identify features that differentiate the two classes Mov: No TC and Mov: TC. For this, we computed the signed SHAP value difference:
\begin{equation*}
\Delta \text{SHAP}_{i,j}
=
\phi_{i,j}^{(\text{Mov: TC})}
-
\phi_{i,j}^{(\text{Mov: No TC})},
\end{equation*}
where $\phi_{i,j}^{(c)}$ denotes the SHAP value of feature $j$ for sample $i$ with respect to class $c$. 
Positive values of $\Delta \text{SHAP}_{i,j}$ indicate that the feature shifts the model prediction towards Mov: TC, whereas negative values indicate a shift toward Mov: No TC, allowing to assess the directionality of each feature contribution. To quantify the overall discriminative strength of each feature independent of its directionality, we introduced a separation score, defined as the mean absolute magnitude of these differences across all samples:
\begin{equation*}
\mathrm{Mean}|\Delta \mathrm{SHAP}_j|
=
\frac{1}{N}
\sum_{i=1}^{N}
\left|
\Delta \text{SHAP}_{i,j}
\right|,
\label{eq:shap_separation}
\end{equation*}
where $N$ denotes the number of samples. Features were subsequently ranked in descending order of this separation score, with larger values indicating stronger class-discriminative contributions. All SHAP analyses were performed on representative subsets sampled from the able-bodied training data using a fixed background dataset. Moreover, to assess feasibility for real-time deployment, the pipeline was implemented using a ring buffer, and the computational latency of its individual components, including pre-processing, feature extraction, and model inference, was evaluated on a standard laptop computer (MacBook Pro, Apple M1 Pro, 16 GB RAM, Apple Inc., USA).

Finally, we evaluated preliminary clinical applicability by testing the final model on the patient dataset using the same evaluation metrics as for the able-bodied cohort. This analysis aimed to assess whether the model, after being validated for subject independent generalization on able-bodied participants, could also generalize across domains simulated impairment to movement patterns of neurological patients. To this end, model performance was first evaluated on patient recordings obtained during reaching tasks performed in a controlled laboratory environment comparable to the able-bodied acquisition protocol. Subsequently, generalization was further assessed on patient data recorded during conventional upper-limb therapy, which involved less constrained movements and increased task variability.

\section{Results}
\label{sec:results}
\subsection{Impact of anatomical locations on CTM detection performance}
To determine the minimal set of anatomical locations capable of providing stable CTM detection, we evaluated the classifier performance across different OMC marker configurations (Fig.~\ref{fig:OMC_locations}; further details in Appendix~\ref{sec:appendix_performance}). Among the different configurations, the three-marker setup (wrist+trunk+upper arm) achieved the highest overall performance across all three classes (F1$_{\text{macro}}$ = 0.86 $\pm$ 0.05, MCC = 0.81 $\pm$ 0.08). Removing the upper-arm marker resulted in only marginal performance changes with comparable spread across folds (F1$_{\text{macro}}$ = 0.85 $\pm$ 0.06, MCC = 0.79 $\pm$ 0.09), with no significant difference in F1$_{\text{Mov: TC}}$ scores between the three-marker and two-marker configurations  ($p_{corr} =  0.625$).%($W = 22$, $p = 0.625$, $r = 0.20$).

In contrast, single-marker configurations using only trunk or wrist data led to a more pronounced decline in classification performance. When relying exclusively on trunk information, performance decreased consistently across all classes (F1$_{\text{macro}}$ = 0.66 $\pm$ 0.12, MCC = 0.50 $\pm$ 0.13), with particularly strong degradation observed for the Calib class (F1$_{\text{Calib}}$= 0.65 $\pm$ 0.16). More detailed analysis of the Mov: TC class further revealed that recall$_{\text{Mov: TC}}$ remained high (wrist+trunk: 0.86 $\pm$ 0.10, trunk-only: 0.83 $\pm$ 0.16), whereas precision$_{\text{Mov: TC}}$ decreased more substantially (wrist+trunk: 0.72 $\pm$ 0.17, trunk-only: 0.60 $\pm$ 0.24), indicating an increased rate of false-positive CTM predictions. Consistent with this performance gap, F1$_{\text{Mov: TC}}$ scores were significantly higher for the two-marker configuration than for the trunk-only configuration ($p_{corr} =  0.012$).%($W = 1$, $p_{\mathrm{Bonf}} = 0.0078$, $r = 0.80$).

For the wrist-only configuration, overall classification performance remained at a moderate level (F1$_{\text{macro}}$ = 0.78 $\pm$ 0.06, MCC = 0.71 $\pm$ 0.08), but performance changes were more class-dependent. While Calib performance remained high (F1$_{\text{Calib}}$ = $0.92 \pm 0.06$), detection performance for the Mov: TC class was reduced, with lower recall$_{\text{Mov: TC}}$ (wrist+trunk: $0.86 \pm 0.10$, wrist-only: $0.77 \pm 0.09$) and precision$_{\text{Mov: TC}}$ (wrist+trunk: $0.72 \pm 0.17$, wrist-only: $0.54 \pm 0.13$). This indicates both a higher number of missed detections and an increased false-positive rate for CTM classification. Accordingly, F1$_{\text{Mov: TC}}$ scores were significantly higher for the two-marker configuration than for the wrist-only configuration ($p_{corr} =  0.012$). For completeness, statistics for all task classes and pairwise model comparisons are reported in the Tab \ref{tab:stats-explo-location} in Appendix.%($W = 1$, $p_{\mathrm{Bonf}} = 0.0078$, $r = 0.80$).

\begin{figure}[t]
    \includegraphics[width=
    \linewidth]{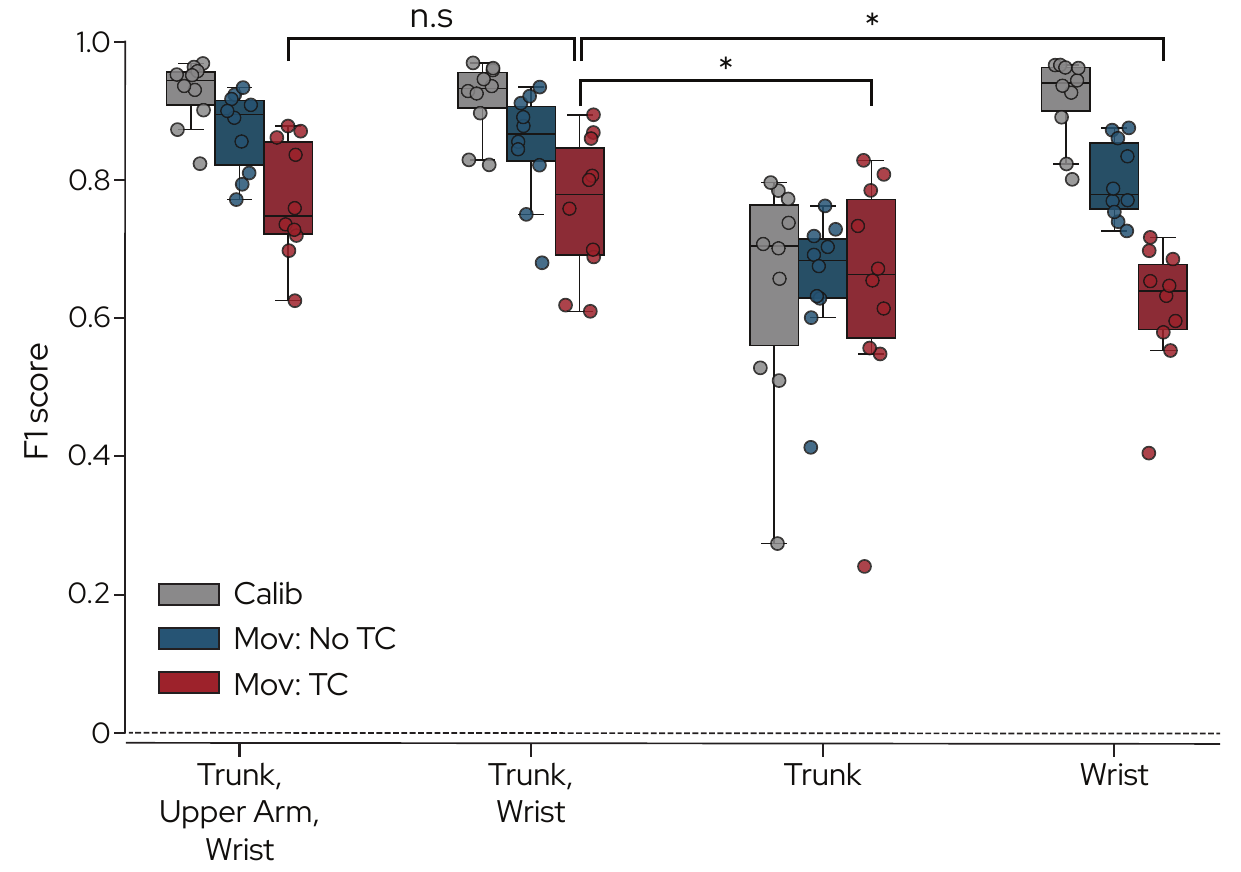}
    \caption{Classification performance across optical motion capture marker configurations in able-bodied participants (n=10). Boxplots show fold-wise F1 scores for the three classes (Calibration, Movement: No Trunk Compensation, and Movement: Trunk Compensation) obtained using leave-one-subject-out cross-validation. Overlaid points indicate individual participant performance. Statistical comparisons on Movement: Trunk Compensation across marker configurations are indicated by brackets (*$p_{corr}<$ 0.05, n.s, not significant). Results for all classes and pairwise model comparisons are reported in Tab \ref{tab:stats-explo-location}.}
    \label{fig:OMC_locations}
\end{figure}

\subsection{Two-IMU classification performance across ADL tasks}

\begin{figure}[t]
    \centering
    \includegraphics[width=\linewidth]{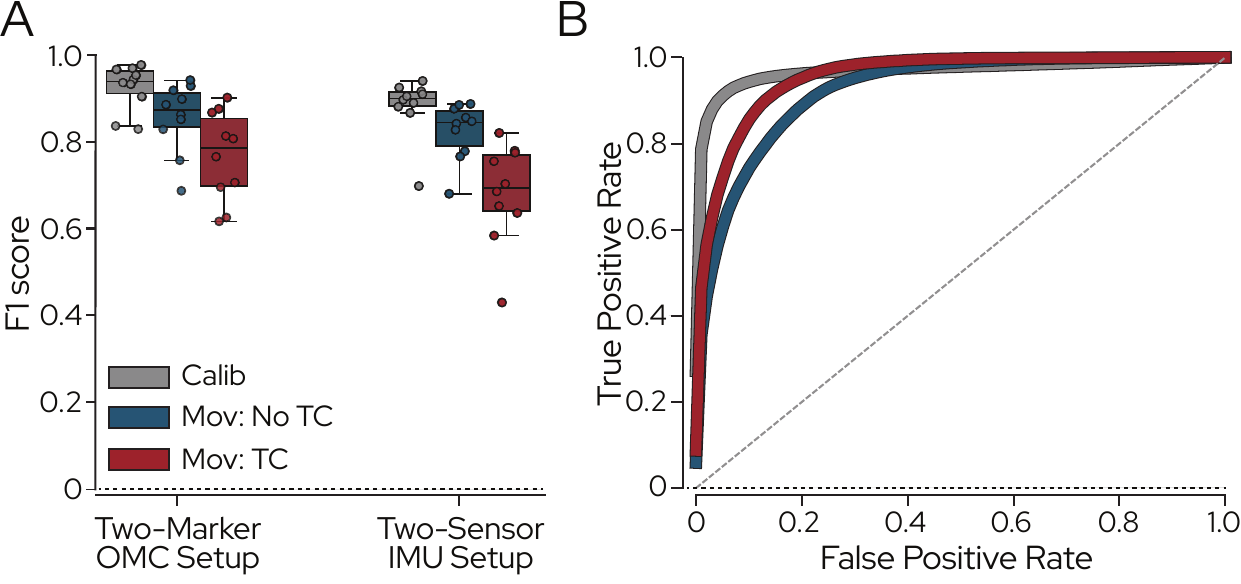} 
    \caption{Two-IMU-based model performance.
    (A) Boxplots show fold-wise F1 scores for optical motion capture (OMC) and inertial measurement unit (IMU) based setups using wrist and trunk signals in able-bodied participants ($n = 10$). Overlaid points indicate individual participant performance. 
    (B) Mean one-vs-rest receiver operating characteristic (ROC) curves across able-bodied participants ($n = 10$) for the three classes demonstrate strong discriminative performance of the IMU model.
    }
    \label{fig:imu_performance}
\end{figure}

% Step 1: General performance assessment
Following the OMC-based analysis, wrist and trunk were used as the minimal sensing locations for subsequent evaluation. The IMU model achieved robust overall performance (F1$_{\text{macro}}$ = 0.80 $\pm$ 0.07, MCC = 0.73 $\pm$ 0.08). While performance was slightly lower than that of the two-marker OMC reference model (F1$_{\text{macro}}$ = 0.85 $\pm$ 0.06, MCC = 0.79 $\pm$ 0.09), inter-subject variability remained comparable between modalities, indicating similar generalization despite reduced sensing complexity (Fig.~\ref{fig:imu_performance}A). Across classes, mean F1 scores were modestly reduced for the IMU configuration ($\Delta$F1$_{\text{Calib}}$ = $-0.04$, $\Delta$F1$_{\text{Mov: No TC}}$ = $-0.03$, $\Delta$F1$_{\text{Mov: TC}}$ = $-0.09$), whereas standard deviations differed by a maximum of 0.015. Statistical comparison using Wilcoxon signed-rank tests with Holm-Bonferroni correction revealed no significant difference between OMC and IMU in the Calib condition (p$_{\text{corr}}$ = 0.064). In contrast, significant differences were observed for both movement conditions, with higher performance for OMC in Mov: No TC (p$_{\text{corr}}$ = 0.020) and Mov: TC (p$_{\text{corr}}$ = 0.012). To further characterize class separability independent of the selected decision rule, we examined one-vs-rest ROC curves averaged across test folds. The resulting curves (Fig.~\ref{fig:imu_performance}B) demonstrated strong discriminative performance across all three classes, as indicated by ROC-AUC values (AUC$_{\text{Calib}}$ = 0.97 $\pm$ 0.05, AUC$_{\text{Mov: No TC}}$ = 0.93 $\pm$ 0.04, AUC$_{\text{Mov: TC}}$ = 0.95 $\pm$ 0.02).

\begin{figure}[t]
       \centering
    \includegraphics[width=\linewidth]{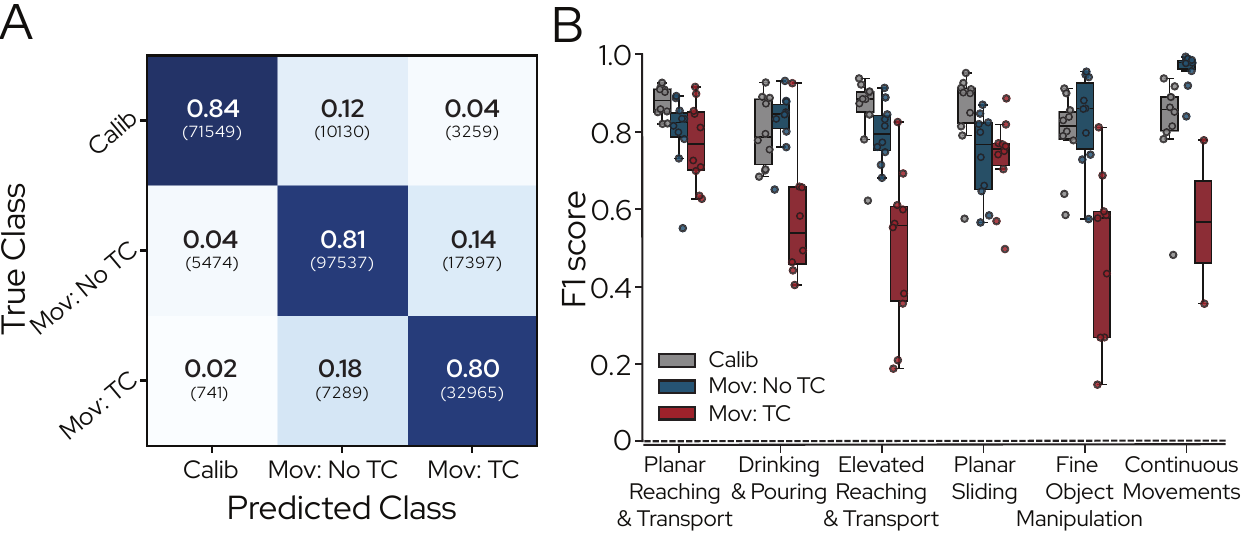} 
    \caption{Detailed analysis of IMU-based model performance. (A) Row-normalized confusion matrix summarizing class-wise prediction outcomes (recall) for Calibration (Calib), Movement: No Trunk Compensation (Mov: No TC), and Movement: Trunk Compensation (Mov: TC), aggregated across all test folds. %Misclassifications predominantly occur between the two movement classes.
    (B) Boxplots depict fold-wise F1 scores across able-bodied participants ($n = 10$), stratified by movement task category. % highlighting task-dependent variability in CTM detection performance.
    Overlaid points indicate individual participant performance.}
    \label{fig: cm_tasks}
\end{figure}

% Step 2: Focus on misclassifications
Further insight into model performance was obtained from the confusion matrix in Fig.~\ref{fig: cm_tasks}A, which was row-normalized to account for class imbalance. Overall, the model achieved class-wise recalls $>$0.8, with most misclassifications occurring between the two movement classes (Mov: No TC vs. Mov: TC). Specifically, 17.8\% of CTMs were misclassified as normal movements, while the reverse confusion occurred in 14.4\% of cases. To characterize misclassification patterns, model performance was stratified by movement task category (Fig.~\ref{fig: cm_tasks}B). Across all categories, Calib and Mov: No TC achieved consistently high mean F1 scores $>$0.73, with low inter-subject variability (std $< 0.13$). In contrast, performance for Mov: TC exhibited pronounced task dependence. While Planar Reaching \& Transport (F1$_{\text{Mov: TC}}$ = 0.77 $\pm$ 0.10) and Planar Sliding (F1$_{\text{Mov: TC}}$ = 0.73 $\pm$ 0.11) achieved stable performance across folds, the remaining task categories showed increased inter-participant variability (Drinking \& Pouring: F1$_{\text{Mov: TC}}$ = 0.58 $\pm$ 0.17, Elevated Reaching \& Transport: F1$_{\text{Mov: TC}}$ = 0.50 $\pm$ 0.21, Fine Object Manipulation: F1$_{\text{Mov: TC}}$ = 0.49 $\pm$ 0.22), with several low-performing folds. In all three of these task categories, reduced F1 scores were primarily driven by decreased recall$_{\text{Mov: TC}}$, indicating that CTMs were frequently misclassified as non-compensatory movements (percentage of CTMs misclassified as non-compensatory movements: Drinking \& Pouring, 25.0\%; Elevated Reaching \& Transport, 41.7\%; Fine Object Manipulation, 39.7\%, Continuous Movements: 22.1 \%).

\begin{figure*}[t]
    \centering
    \includegraphics[width=0.85\textwidth]{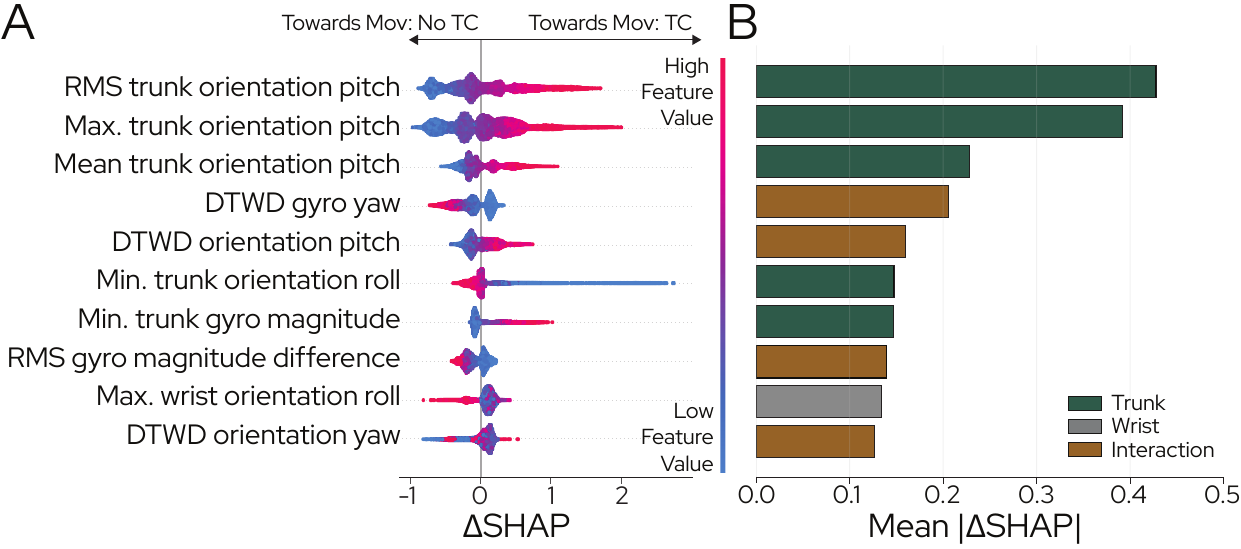}
    \caption{Explainability analysis using Shapley Additive Explanations (SHAP) to identify the top 10 discriminative features between Movement: No Trunk Compensation (Mov: No TC) and Movement: Trunk Compensation (Mov: TC).
    (A) Distributions of signed $\Delta$SHAP values indicating directional class support, where positive values favor Mov: TC and negative values favor Mov: No TC. Feature values are color-coded from low (blue) to high (red) based on their relative magnitude.
    (B) Separation score expressed as Mean$|\Delta$SHAP$|$ per feature summarizing overall discriminative importance. Bars are color-coded by feature origin (trunk, wrist, interaction).}
    \label{fig:shap}
\end{figure*}

To identify the most discriminative kinematic features separating Mov: No TC and Mov: TC, SHAP-based model explainability was applied to the final IMU-based model. Features were ranked according to their separation score (Fig.~\ref{fig:shap}). To assess the relative contribution of different feature groups (wrist, trunk, interaction), the separation scores $ \text{Mean}|\Delta \text{SHAP}|$ were aggregated within each group. Within the top-10 predictors, trunk-derived features accounted for 64\% of the total separation score, wrist--trunk interaction features for 30\%, and isolated wrist features for 6\%. These proportions indicate that the model’s decision boundary was predominantly shaped by trunk kinematics, while coordination-based interaction features provided complementary discriminative evidence.

In terms of directional effects, increased trunk flexion (orientation pitch) and overall trunk movement intensity (gyroscope magnitude) as well as altered trunk side lean (orientation roll) were consistently associated with positive $\Delta$SHAP values, indicating increased model confidence towards the Mov: TC class. Wrist–trunk interaction features describing relative segment orientation (Dynamic Time Warping Distance, DTWD, of orientation pitch/yaw) and movement (DTWD gyro yaw and gyro magnitude difference) showed distinct patterns in their $\Delta$SHAP distributions. Indeed, movement-related interaction features exhibited a clear directional tendency, with lower relative values (i.e. more similar movement of trunk and wrist) predominantly associated with positive $\Delta$SHAP contributions toward Mov: TC. In contrast, orientation-based interaction features displayed more heterogeneous, bidirectional distributions, with both positive and negative contributions observed across the feature value range.

Real-time suitability of the pipeline was assessed by measuring the computational latency of its individual components over $n=5710$ samples. Model inference required $29.25 \pm 10.93$~ms, while pre-processing and feature extraction required $42.47 \pm 17.31$~ms and $46.86 \pm 16.68$~ms, respectively.

\subsection{Patient evaluation}
\begin{figure*}[t]
    \centering
    \includegraphics[width=0.85\textwidth]{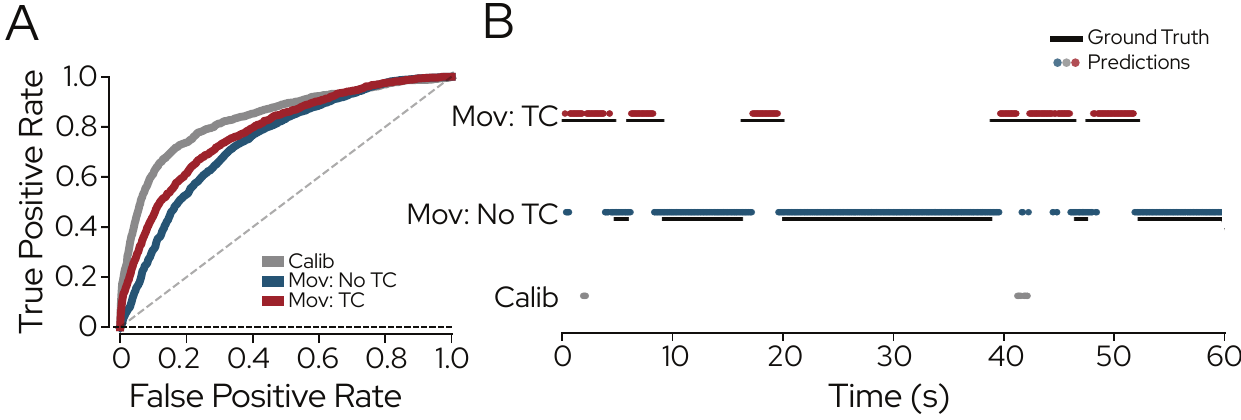}    

    \caption{Detection performance with clinical dataset. (A) Mean one-vs-rest receiver operating characteristic (ROC) curves across across patients performing tasks with the most affected arm, with wrist sensor located on the most affected limb ($n=4$) (B) Representative 60 s time-series excerpt showing ground-truth and predicted class labels for a patient performing conventional upper-limb therapy. Ground-truth annotations are shown as a black line, and model predictions as colored dots for the three classes (Movement: Trunk Compensation, Movement: No Trunk Compensation, and Calibration).}%(B) Sensor is placed on the less affected limb ($n=3$).}
    \label{fig:patients_roc}
\end{figure*}

To assess preliminary clinical generalization, the final model trained exclusively on able-bodied recordings under simulated impairment was evaluated on data from four patients.
Across these recordings, the model retained strong discriminative capability, with AUC$_{\text{macro}}$ values of 0.78 $\pm$ 0.06 (Fig. \ref{fig:patients_roc}A, details in Table~\ref{tab:patient_affected_results}). In particular, AUC$_{\text{Mov: No TC}}$ remained high (0.78 $\pm$ 0.10), indicating that compensatory and non-compensatory trunk movement segments remained separable in patient recording.%ranging from 0.72 to 0.85 

In contrast, threshold-dependent classification metrics showed moderate performance, with F1$_{\text{macro}}$ = 0.43 $\pm$ 0.06 and MCC = 0.28 $\pm$ 0.04. Movement phases were detected most consistently (F1$_{\text{Mov: No TC}}$ =
$0.65 \pm 0.16$), whereas compensation detection varied substantially across patients
(precision$_{\text{Mov: TC}}$ = $0.57 \pm 0.27$). For example, P63 achieved high precision$_{\text{Mov: TC}}$ (0.96), but low recall$_{\text{Mov: TC}}$ (0.11), indicating that predicted CTM events were frequently correct, but that a substantial portion of CTM time points was missed. This trend was also observed in the conventional therapy recording of P03, a more challenging and less constrained evaluation scenario. Here, the model achieved high performance for Mov: No TC (precision$_{\text{Mov: No TC}}$ = 0.898, recall$_{\text{Mov: No TC}}$ = 0.963, F1$_{\text{Mov: No TC}}$ = 0.929), while CTM detection again showed relatively high precision$_{\text{Mov: TC}}$ (0.674), but low recall$_{\text{Mov: TC}}$ (0.266), resulting in an F1$_{\text{Mov: TC}}$ score of 0.381. A representative time-series segment of this recording is shown in Fig.~\ref{fig:patients_roc}B, illustrating the temporal evolution of model predictions relative to the ground-truth annotations.

\section{Discussion}
\label{sec:discussion}

In this study, we evaluated the feasibility of ML–based detection of CTMs using a minimal IMU configuration. Specifically, we hypothesized that trunk and wrist kinematics alone would provide sufficient information for robust, real-time CTM detection during ADL tasks in unconstrained workspaces. We further assessed whether this reduced sensor setup generalizes from able-bodied recordings performed under simulated impairment to preliminary clinical data acquired in laboratory and conventional upper-limb therapy settings.

\subsection{Wrist and trunk provide complementary information}

Regarding the impact of sensor placement on CTM detection performance, both the OMC-based marker removal study and IMU-based SHAP feature analysis yielded consistent evidence that the most discriminative information is captured together by trunk and wrist kinematics. The two-marker configuration (wrist+trunk) achieved performance comparable to a three-marker setup (wrist+trunk+upper arm), whereas single-location models showed significantly reduced performance, with the trunk-only model exhibiting decreased precision due to increased false-positive CTM predictions. Feature-level explainability further highlighted dominant trunk-derived features as well as high-ranking wrist–trunk interaction features, indicating that relative coordination patterns contribute substantially to CTM discrimination.

These findings align with prior biomechanical work describing compensation as altered trunk–distal coordination \cite{cirstea_2000, brami_2021} and with IMU-based studies reporting strong performance for wrist–trunk configurations compared to trunk-only sensing \cite{ranganathan_2017}. While trunk kinematics constitute a primary indicator of CTMs, the reduced precision of the trunk-only model suggests limited ability to distinguish compensatory from physiologically appropriate trunk involvement. Incorporating wrist signals improves discrimination by capturing coordination and orientation cues, as observed in the SHAP analysis. Moreover, the limited incremental benefit of upper-arm sensing for this classification task, despite its importance for detailed biomechanical analyses \cite{schwarz_2020, schwarz_2021}, may be explained by redundancy with wrist-derived information. Overall, the wrist–trunk configuration represents a practical trade-off between performance and deployability, reducing instrumentation burden and calibration effort for unsupervised daily-life use.

\noindent
\subsection{ML-based CTM detection achieves robust performance but remains task-dependent}
Our proposed two-IMU model achieved strong discriminative performance for CTM detection, reflected by high ROC-AUC values as well as stable macro-averaged F1 and MCC scores across able-bodied participants. Although performance was slightly lower compared to the two-marker OMC reference model, inter-subject variability remained comparable, indicating similar generalization despite reduced sensing complexity.  

Previous studies have demonstrated the feasibility of CTM detection across diverse sensing modalities and classification approaches. IMU-based systems have reported high performance (often $>$90\% accuracy) using trial- or repetition-level classification on pre-segmented data \cite{ding_2024, ranganathan_2017, ranganathan_2017_2}. However, these approaches do not provide continuous predictions during task execution, limiting their applicability for real-time monitoring in daily-life settings. Continuous CTM detection has instead primarily been achieved using camera- \cite{lin_2023} or pressure-based systems \cite{cai_2020, cai_2020_2}, which offer dense temporal information but rely on less portable sensing setups. 

Task-dependent performance differences were observed in the able-bodied dataset, with increased misclassifications between Mov: No TC and Mov: TC particularly for Drinking and Pouring, Elevated Reaching and Transport, and Fine Object Manipulation. These errors were mainly driven by missed CTM events, reducing recall, consistent with prior reports \cite{ranganathan_2017, seo_2024}. This may be explained by weaker and more heterogeneous compensatory patterns across participants, as well as labeling ambiguities in tasks that do not reliably elicit pronounced CTMs \cite{sauerzopf_2024}. Moreover, window-based inference is sensitive to ambiguous onset and offset transitions, which may further increase misclassification rates.

\noindent
\subsection{Generalization to patient data is feasible but variable}
Evaluation on patient data provided initial evidence that compensation-related kinematic signatures learned from able-bodied recordings under simulated impairment are, to some extent, transferable to real post-stroke movement patterns. Across four recordings, consistently high ROC-AUC values indicated that the model retained strong discriminative capability despite the increased heterogeneity and reduced repeatability of patient movements. 

In contrast, threshold-dependent metrics such as macro-averaged F1 and MCC were only moderate, consistent with prior reports highlighting the challenges of robust compensation labeling \cite{sauerzopf_2024} and classification \cite{unger_2025} across heterogeneous neurological populations. The mismatch between ROC-AUC and threshold-dependent metrics suggests that the model preserved discriminative information, but that differences in patient movement strategies, class prevalence, and temporal ambiguity at movement onset and offset affected the optimal decision threshold and increased classification errors. In particular, the observed combination of relatively high precision but low recall for the CTM class, similar to what was observed in several able-bodied tasks, suggests a conservative prediction behavior. This conservative prediction behavior may be beneficial, since false-positive CTM detection could incorrectly penalize normal movements, whereas missed compensation time points mainly reduce sensitivity. Importantly, the therapy recording (P03) represented a more realistic and challenging out-of-laboratory evaluation scenario. Despite these conditions, the model maintained strong movement-phase detection and moderate CTM detection performance, supporting the feasibility of real-time deployment.

\noindent
\subsection{Limitations and future work} 

A primary limitation of this study is that the proposed ML model was developed using a small cohort of able-bodied participants, in whom CTMs were induced through externally imposed movement constraints. While such paradigms have been used in prior work \cite{ranganathan_2017, berjis_2025}, compensatory strategies following stroke emerge from underlying neurological impairments and are therefore expected to exhibit more complex and heterogeneous kinematic signatures and inter-joint coordination patterns \cite{cirstea_2000, levin_2009}. Although the applied brace and resistance-band constraints were motivated by movement limitations commonly reported in stroke survivors \cite{beer_1999,mccrea_2002,dewald_1995,ellis_2017}, they likely capture only a subset of clinically observed compensation behaviors, particularly across varying impairment severity. As a consequence, models trained on able-bodied data may learn decision rules that only partially generalize to real post-stroke movement patterns, which is reflected in the moderate and variable performance observed in the patient evaluation. Future work should therefore include larger and more heterogeneous patient cohorts spanning a broad range of impairment severities to enable model training and validation directly on clinical movement variability. At the same time, able-bodied recordings under simulated impairment may still serve as an efficient strategy for data augmentation, particularly when large-scale patient recordings are limited. In addition, prediction probabilities could be exploited to temporally aggregate frame-wise outputs into more robust movement-level decisions in real-time. Class-specific probability thresholds may further allow application-dependent tuning of false-positive and false-negative CTM detections, potentially improving usability for feedback-oriented deployment.

A second limitation is the use of a coarse categorical labeling scheme indicating only the presence or absence of CTMs. Grouping subtle and pronounced compensations into a single class may obscure clinically relevant differences and contribute to misclassifications at class boundaries. Employing graded compensation severity labels (e.g., 0--3), for example based on the RPS~\cite{levin_2004}, could enable more fine-grained evaluation of model behavior and reveal whether detection performance varies systematically with CTM severity. Such stratification would be particularly relevant for feedback applications, where timely intervention is most critical in the presence of pronounced compensatory behavior.

Despite these limitations, the proposed approach highlights the potential of minimal wearable sensing for continuous CTM monitoring. Given its real-time capability, the system could be directly extended to provide feedback, for example by combining a trunk-mounted wearable (e.g., necklace-style) with a wrist-worn device, enabling patients to receive auditory or visual cues when CTMs are detected during supervised therapy and unsupervised home training. This may extend movement-quality feedback beyond the clinic, reduce therapist workload, and support more consistent and objective feedback across settings. Moreover, continuous monitoring over longer time scales could enable objective tracking of compensation frequency and duration during daily activities, potentially supporting clinical decision-making and longitudinal assessment of rehabilitation progress.

\section{Conclusion}
\label{sec:conclusion}

This study investigated the feasibility of machine learning--based detection of CTMs using a minimal wearable sensing configuration. Using OMC recordings from able-bodied participants, mimicking post-stroke movement limitations in upper-limb tasks, we identified wrist and trunk kinematics as the most informative anatomical locations for CTM detection. Building upon this analysis, we developed a two-IMU (trunk--wrist) model capable of continuous real-time inference, achieving robust and stable classification performance across able-bodied subjects. When applied to preliminary patient recordings acquired during reaching tasks and conventional upper-limb therapy, the model retained high discriminative capability but showed reduced and variable threshold-dependent performance, highlighting the challenges of generalization to heterogeneous clinical movement patterns. Future work will focus on retraining and validation with larger and more diverse patient cohorts and on integrating the proposed approach into closed-loop rehabilitation and assistive systems.

\section*{Acknowledgment} 
The authors thank Abigail Vogel for her contributions to figure design and to the fabrication of wearable hardware for sensor and marker attachment. In addition, we would like to thank Alen Roady for his essential contribution to the data acquisition software. We further thank Manuel Ehrler for his work on the real-time inference pipeline and Sho Sugimoto for his support with preliminary evaluation of patient data. We are also grateful to Jeannine Müller for her support and collaboration during clinical data collection.

\section*{Conflict of Interest} 
MQ, OL, PV and DD are affiliated with Skaaltec AG, which is involved in the development/commercialization of technology described in this study. The remaining authors declare no competing interests.

\appendices

\section{Workspace, Task Setup and Movement Tasks}
\label{sec:appendix_workspace}
\begin{figure*}[!htbp]
  \centering
  \includegraphics[width=\textwidth]{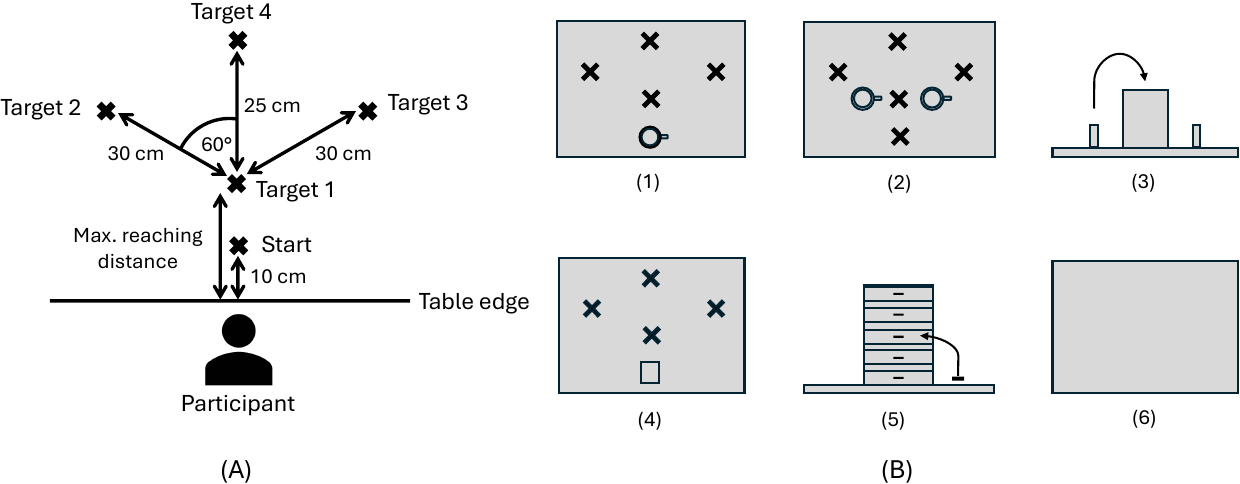}
  \caption{Experimental workspace and task setup. (a) Schematic of the workspace showing the fixed start position (10~cm from the table edge) and four target locations. Target~1 (T1) corresponds to the participant’s maximum reaching distance, while Targets~2--4 (T2--T4) are placed at predefined radial distances (25--30~cm) with 60$^\circ$ angular separation. (b) Overview of task categories: (1) planar reaching and transport; (2) drinking and pouring; (3) elevated reaching and transport; (4) planar sliding; (5) fine object manipulation; and (6) continuous movements.}
  \label{fig:workspace}
\end{figure*}

This appendix provides additional methodological details and results that complement the main manuscript. The experimental workspace and task setup are illustrated in Fig.~\ref{fig:workspace}, while the corresponding movement tasks are summarized in Tab.~\ref{tab:tasks}.

\begin{table*}[!t]
\centering
\footnotesize
\renewcommand{\arraystretch}{1.05}
\setlength{\tabcolsep}{5pt}

\begin{tabularx}{\textwidth}{
    >{\raggedright\arraybackslash}p{0.18\textwidth}
    >{\raggedright\arraybackslash}p{0.22\textwidth}
    >{\raggedright\arraybackslash}X
}
\toprule
\textbf{Task Category} & \textbf{Objects \& Setup} & \textbf{Movement Tasks} \\
\midrule

Planar Reaching \& Transport
& Single cup at Start
& \begin{tabular}[t]{@{}l@{}}
-- Reaching to T1--T4 (Task No.~1--4) \\
-- Cup transport between Start and T1--T4 (Task No.~5--8) \\
-- Sequential transport: Start $\rightarrow$ T2 $\rightarrow$ T3 $\rightarrow$ Start \\
\quad and Start $\rightarrow$ T3 $\rightarrow$ T2 $\rightarrow$ Start (Task No.~9--10) \\
-- Continuous reaching between T1--T4 in pseudo-random order (Task No.~11--12)
\end{tabular}
\\

\addlinespace[6pt]

Drinking \& Pouring
& Two cups positioned to the left and right of T1
& \begin{tabular}[t]{@{}l@{}}
-- Water pouring (right $\leftrightarrow$ left) (Task No.~13--14) \\
-- Drinking from right and left cup (Task No.~15--16)
\end{tabular}
\\

\addlinespace[6pt]

Elevated Reaching \& Transport
& One large block at T1 and two smaller blocks on either side
& \begin{tabular}[t]{@{}l@{}}
-- Elevation of small block onto large block with $180^\circ$ flip and return (Task No.~17--18) \\
-- Transport over large block to opposite side and return (Task No.~19--20)
\end{tabular}
\\

\addlinespace[6pt]

Planar Sliding
& Single card at Start
& \begin{tabular}[t]{@{}l@{}}
-- Two-handed sliding: Start $\rightarrow$ T2 $\rightarrow$ T3 $\rightarrow$ Start \\
\quad and Start $\rightarrow$ T3 $\rightarrow$ T2 $\rightarrow$ Start (Task No.~21--22) \\
-- Two-handed sliding: Start $\rightarrow$ T1 $\rightarrow$ Start and Start $\rightarrow$ T4 $\rightarrow$ Start (Task No.~23--24) \\
-- Single-handed sliding to T1--T4 with $180^\circ$ card flip (Task No.~25--28) \\
-- Single-handed sequences: Start $\rightarrow$ T2 $\rightarrow$ T3 $\rightarrow$ Start \\
\quad and Start $\rightarrow$ T3 $\rightarrow$ T2 $\rightarrow$ Start (Task No.~29--30) \\
-- Continuous multi-target sliding sequences (Task No.~31--32)
\end{tabular}
\\

\addlinespace[6pt]
Fine Object Manipulation
& Five-drawer box at T1 and a coin placed to its right
& \begin{tabular}[t]{@{}l@{}}
-- Coin manipulation across drawers (opening, closing, insertion, transfer) (Task No.~33--34) \\
-- Sequential opening and closing of all drawers (Task No.~35--36)
\end{tabular}
\\

\addlinespace[6pt]

Continuous Movements
& ---
& \begin{tabular}[t]{@{}l@{}}
-- Hand-biking motion (20--30~s) (Task No.~37) \\
-- Swimming motion (20--30~s) (Task No.~38)
\end{tabular}
\\

\bottomrule
\end{tabularx}

\caption{Overview of task categories, objects, and movement tasks defined relative to target locations T1--T4 in the data acquisition protocol for able-bodied participants.}
\label{tab:tasks}
\end{table*}

\section{Labeling Guidelines}
\label{sec:appendix_labeling}
Labeling guidelines are presented in Tab. \ref{tab:labeling_guidelines}.

\begin{table*}[!htbp]
\centering
\renewcommand{\arraystretch}{1.2}
\begin{tabular}{p{4cm} p{11cm}}
\toprule
\textbf{Label} & \textbf{Description} \\
\midrule

\textbf{Movement: Trunk Compensation} &
\textbf{Movement Pattern:} Visible trunk displacement in the sagittal plane relative to the pelvis. Elbow extension and shoulder flexion remain below the required range, and trunk flexion compensates for limited reach. \newline
\textbf{Onset:} Initiation of forward trunk displacement with limited shoulder and elbow motion. \newline
\textbf{Offset:} Return to upright posture and stabilization. \\

\textbf{Movement: No Trunk Compensation} &
\textbf{Movement Pattern:} Arm movement occurs while trunk displacement remains within the functional range required for task execution. No compensatory trunk motion is observed. \newline
\textbf{Onset:} Visible arm movement while the trunk remains stable. \newline
\textbf{Offset:} Arm returns to calibration position and posture stabilizes. \\

\textbf{Calibration} &
\textbf{Movement Pattern:} Arm rests in a bent position close to the body with no visible trunk or wrist movement. Neutral alignment of shoulders, pelvis, and spine. \newline
\textbf{Onset:} Completion of previous task and stabilization in calibration posture. \newline
\textbf{Offset:} Initiation of movement via visible displacement of wrist, shoulder, or trunk. \\

\bottomrule
\end{tabular}
\caption{Labeling guidelines for the three movement classes, including movement characteristics and onset/offset definitions.}
\label{tab:labeling_guidelines}
\end{table*}

\section{Anatomical Calibration and Left Arm Transformation}
\label{sec:appendix_calibration}
\begin{figure}[!t]
    \centering
    \includegraphics[width=\columnwidth]{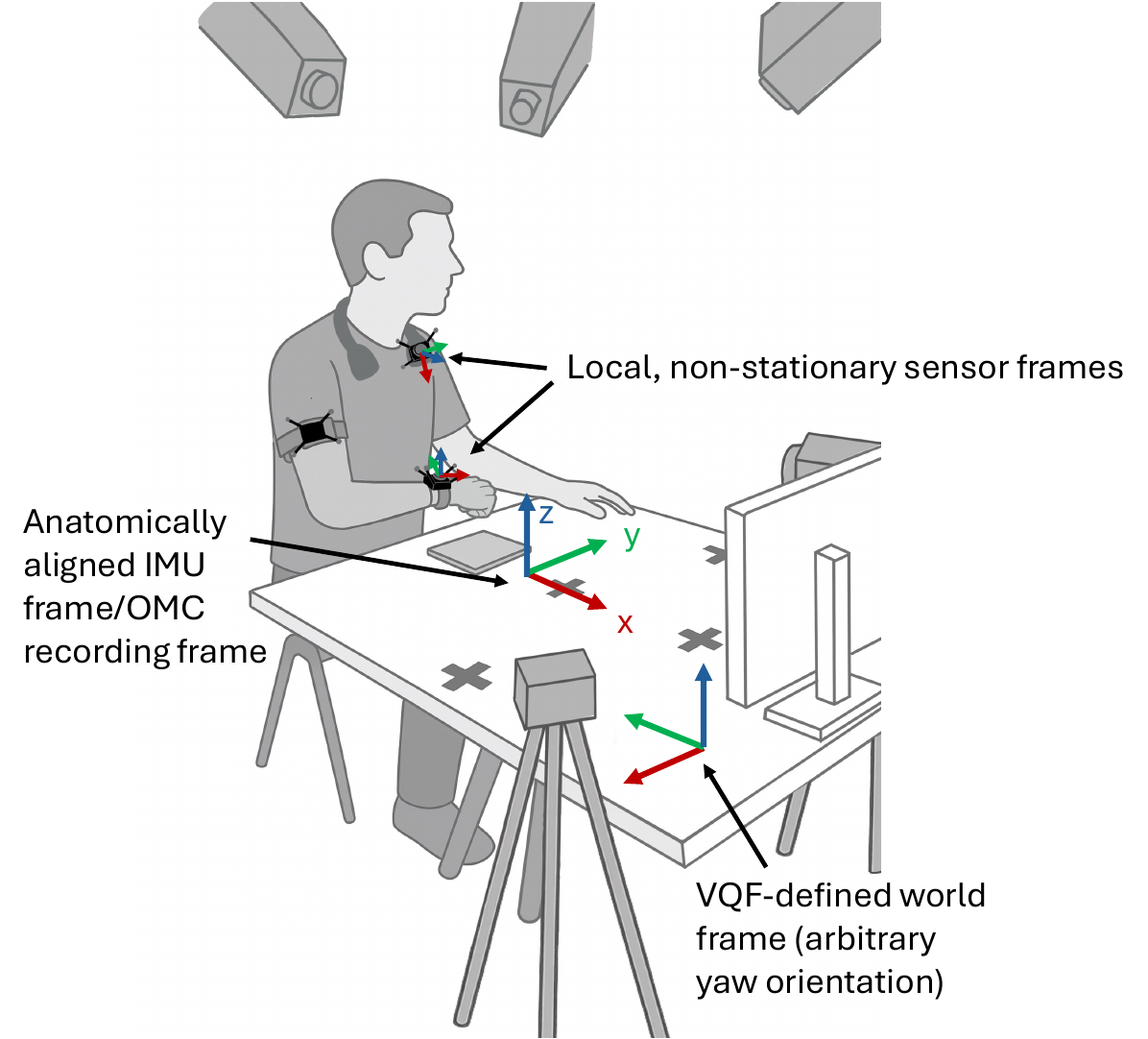}
    \caption{Coordinate frames used for IMU-based anatomical calibration: Local sensor frames are attached to each IMU and vary with segment motion, while the VQF estimates orientations in a global world frame with arbitrary yaw. A standardized calibration pose is used to define a sensor-specific anatomically aligned frame that matches the OMC recording frame, allowing consistent interpretation of sensor orientations.}
    \label{fig:anatomical_calibration}
\end{figure}

Following the approach of Nguyen et al.~\cite{nguyen_2021}, anatomical calibration was used to express all IMU-based kinematics in an anatomically aligned coordinate frame (Fig.\ref{fig:anatomical_calibration}). This anatomical frame was initialized from the first sample of each recording while participants held the standardized calibration pose introduced in Sec.~\ref{sec:acquisition}. To establish this frame, we first extracted the world-frame orientation quaternion $\mathbf{q}_{\text{world},s,0}$ at the first sample for each sensor $s \in \{\text{wrist}, \text{trunk}\}$ as estimated by the Versatile Quaternion Filter (VQF). In a second step, we assumed that during the calibration pose one local sensor axis aligns with the anatomical left--lateral direction. Due to the sensor mounting, this corresponds to the local $x$-axis for the wrist sensor and the local $y$-axis for the trunk sensor. The corresponding world-frame directions were obtained by rotating the respective pure quaternion axis vectors using the initial sensor orientation:
\begin{align*}
\mathbf{q}_{\text{lat,wrist}}
&= \mathbf{q}_{\text{world,wrist},0}
\;\otimes\; [0,1,0,0]^\top
\;\otimes\; \mathbf{q}_{\text{world,wrist},0}^{-1} \\
\mathbf{q}_{\text{lat,trunk}}
&= \mathbf{q}_{\text{world,trunk},0}
\;\otimes\; [0,0,1,0]^\top
\;\otimes\; \mathbf{q}_{\text{world,trunk},0}^{-1}
\end{align*}
Dropping the real part of $\mathbf{q}_{\text{lat},s}$ yields the corresponding 3D direction vectors, denoted as
$\tilde{\mathbf{y}}_{\text{calib},s} \in \mathbb{R}^3$. For both sensors, the vertical axis of the calibrated frame was fixed as
$\mathbf{z}_{\text{calib}} = [0,0,1]^\top$ (world up direction).
A right-handed orthonormal basis was constructed via cross products:
\begin{align*}
\tilde{\mathbf{x}}_{\text{calib},s} &= 
\tilde{\mathbf{y}}_{\text{calib},s} \times \mathbf{z}_{\text{calib}}, \\
\mathbf{x}_{\text{calib},s} &=
\frac{\tilde{\mathbf{x}}_{\text{calib},s}}
{\lVert \tilde{\mathbf{x}}_{\text{calib},s} \rVert}, \\
\mathbf{y}_{\text{calib},s} &=
\mathbf{z}_{\text{calib}} \times \mathbf{x}_{\text{calib},s},
\qquad s \in \{\text{wrist}, \text{trunk}\}.
\end{align*}

\noindent
Stacking the unit axes yields the sensor-specific rotation matrix
\begin{equation*}
\mathbf{R}_{\text{rot},s}
=
\big[
\mathbf{x}_{\text{calib},s}\;\;
\mathbf{y}_{\text{calib},s}\;\;
\mathbf{z}_{\text{calib}}
\big],
\end{equation*}
and the equivalent quaternion $\mathbf{q}_{\text{rot},s}$, representing the rotation from the calibrated anatomical frame to the VQF world frame. Any sensor orientation $\mathbf{q}_{\text{world},s}(t)$ expressed in the VQF world frame was then mapped into the calibrated frame by
\begin{equation*}
\mathbf{q}_{\text{calib},s}(t)
=
\mathbf{q}_{\text{rot},s}^{-1}
\;\otimes\;
\mathbf{q}_{\text{world},s}(t),
\qquad s \in \{\text{wrist}, \text{trunk}\}.
\end{equation*}

\noindent
For patient recordings in which the calibration pose was performed with the arm resting on the table and aligned with the sagittal plane, the resulting anatomical frame differed by a fixed in-plane rotation. To account for this, the calibrated coordinate frame was rotated by 90° about the vertical ($z$) axis, ensuring consistency with the anatomical frame definition used for able-bodied recordings. This adjustment preserved the alignment of anatomical axes while accommodating the modified calibration posture:
\begin{equation*}
\mathbf{q}_{z} =
\begin{bmatrix}
\cos(\theta/2) & 0 & 0 & \sin(\theta/2)
\end{bmatrix}^\top,
\qquad \theta = 90^\circ,
\end{equation*}

\noindent
Such that the corrected orientation was obtained as:

\begin{equation*}
\mathbf{q}_{\text{calib},wrist}^\prime(t)
=
\mathbf{q}_{z}
\;\otimes\;
\mathbf{q}_{\text{calib},wrist}(t)
\end{equation*}

\noindent
For patient recordings acquired from the left arm, kinematic data were mirrored with respect to the sagittal plane to obtain right-arm–equivalent representations. This step was applied to ensure consistency across subjects and to allow all recordings to be processed using a unified anatomical reference.

Mirroring was implemented by reflecting the sensor signals across the sagittal plane using the reflection matrix
\begin{equation*}
\mathbf{R}_{\text{mirror}} =
\begin{bmatrix}
1 & 0 & 0 \\
0 & -1 & 0 \\
0 & 0 & 1
\end{bmatrix}.
\end{equation*}
Linear acceleration signals were transformed as
\begin{equation*}
\mathbf{a}_{\text{mirrored}}(t)
=
\mathbf{a}(t)\,\mathbf{R}_{\text{mirror}}^\top,
\end{equation*}
whereas angular velocity signals were additionally sign-inverted to account for the pseudovector nature of rotational quantities:
\begin{equation*}
\boldsymbol{\omega}_{\text{mirrored}}(t)
=
-\,\boldsymbol{\omega}(t)\,\mathbf{R}_{\text{mirror}}^\top.
\end{equation*}

This transformation preserves the physical consistency of the kinematic signals while mapping left-arm motion into a right-arm–equivalent coordinate representation.

Furthermore, for some patient recordings, the trunk IMU was mounted on the sternum using a textile-based elastic fixation system. Due to variability in sensor placement across recording sessions, the initial sensor orientation relative to the trunk could differ from the reference configuration used in the able-bodied dataset.

To compensate for this offset, an additional alignment step was applied to the calibrated trunk orientation quaternions $\mathbf{q}_{\text{calib},\text{trunk}}(t)$. Specifically, a reference quaternion $\mathbf{q}_{\text{ref}}$ was defined based on the average initial trunk orientation observed in the able-bodied recordings. Let $\mathbf{q}_{\text{calib},\text{trunk}}(0)$ denote the first calibrated trunk quaternion of a given patient recording. If the difference between $\mathbf{q}_{\text{calib},\text{trunk}}(0)$ and $\mathbf{q}_{\text{ref}}$ exceeded a predefined threshold (norm of 0.1), a correction quaternion was computed as
\begin{equation*}
\mathbf{q}_{\text{corr}}
=
\mathbf{q}_{\text{ref}}
\;\otimes\;
\mathbf{q}_{\text{calib},\text{trunk}}(0)^{-1}.
\end{equation*}
The aligned trunk orientation trajectory was then obtained by left-multiplying all
calibrated trunk quaternions by the correction quaternion:
\begin{equation*}
\mathbf{q}_{\text{calib},\text{trunk}}^{\prime}(t)
=
\mathbf{q}_{\text{corr}}
\;\otimes\;
\mathbf{q}_{\text{calib},\text{trunk}}(t).
\end{equation*}

This procedure ensured consistent initial alignment of the trunk IMU coordinate frame across recordings, thereby reducing variability introduced by session-specific
sternum sensor placement.

\section{Kinematic Data Streams and Statistical Features}
\label{sec: kin_streams}

We refer to the native IMU sensor frame as the \textit{local frame} and to the anatomically aligned frame obtained through calibration (see Appendix \ref{sec:appendix_calibration}) as the \textit{calibrated frame}. The OMC reference frame, configured to match the anatomically aligned IMU frame and enabling direct cross-modal comparison of kinematic variables, is referred to as the \textit{recording frame}. Overview of the kinematic data streams is presented in Tab.~\ref{tab:kinematic_streams}, and summary of feature category in Tab.~\ref{tab:feature_metrics}.

\begin{table*}[!htbp]
\centering

\begin{tabularx}{\textwidth}{@{}p{1.7cm} X p{2.2cm}@{}}
\toprule
\textbf{Modality} & \textbf{Kinematic data streams} & \textbf{Frame} \\
\midrule
IMU &
Wrist/trunk acceleration (3-axis \& magnitude) &
Local \& Calibrated \\
&
Wrist/trunk angular velocity (3-axis \& magnitude) &
Local \& Calibrated \\
&
Absolute wrist/trunk orientation (3-axis) &
Calibrated \\
&
Relative wrist--trunk orientation (3-axis \& 3D rotation angle) &
Calibrated \\
\midrule
OMC &
Wrist/trunk/upper-arm acceleration (3-axis \& magnitude) &
Recording \\
&
Wrist/trunk/upper-arm angular velocity (3-axis \& magnitude) &
Recording \\
&
Wrist/trunk/upper-arm linear velocity (3-axis \& magnitude) &
Recording \\
&
Absolute wrist/trunk/upper-arm orientation (3-axis) &
Recording \\
&
Relative wrist--trunk/wrist--upper arm/upper arm--trunk orientation (3-axis \& 3D rotation angle) &
Recording \\
&
Relative wrist--trunk/wrist--upper arm/upper arm--trunk distance (3-axis \& magnitude) &
Recording \\
\bottomrule
\end{tabularx}

\caption{Summary of sensing modalities and available kinematic data streams after pre-processing, together with their respective coordinate frames. Three-axis quantities are represented along the Cartesian axes (x/y/z) or rotational axes (roll/pitch/yaw). Magnitude denotes the Euclidean norm of the corresponding vector signal. The 3D rotation angle represents the total angular displacement between two orientations.}
\label{tab:kinematic_streams}
\end{table*}

\begin{table*}[!htbp]
\centering
\small
\renewcommand{\arraystretch}{1.2}
\setlength{\tabcolsep}{6pt}
\begin{tabularx}{\textwidth}{@{}p{6.2cm} >{\raggedright\arraybackslash}X@{}}
\toprule
\textbf{Metric Category} & \textbf{Measures} \\
\midrule
Univariate channel-wise statistics &
Mean, standard deviation (std), minimum (min), maximum (max), range, root mean square (RMS), mean absolute deviation (MAD), trend, skewness, kurtosis \\
\addlinespace
Bivariate channel-wise similarity metrics (applied on acceleration, absolute orientation, linear/angular velocity) &
Maximum normalized cross-correlation, normalized dynamic time warping distance (DTWD) \\
\addlinespace
Smoothness metrics (applied on acceleration and linear velocity) &
Logarithmic dimensionless jerk (LDJ) for wrist \& trunk + LDJ ratio between wrist \& trunk \\
\bottomrule
\end{tabularx}
\caption{Summary of feature metric categories applied to the kinematic data streams and derived descriptors listed in Tab.~\ref{tab:kinematic_streams}. Univariate metrics are computed independently for each input signal, whereas bivariate metrics quantify relationships between pairs of signals. Channel-wise metrics are computed separately for each signal component (e.g., x/y/z for vector quantities or roll/pitch/yaw for Euler angles), with each component treated as an individual channel (i.e., a single dimension of a multi-dimensional signal) within a kinematic data stream. For 3D vector signals, the Euclidean norm is additionally treated as an extra channel. All metrics are computed within each analysis window of the sliding-window segmentation and serve as input features for the classification model.}
\label{tab:feature_metrics}
\end{table*}

\section{Able-Bodied Model Evaluation}
\label{sec:appendix_performance}
Performance classification results for able-bodied participants using OMC and IMU are presented in Tab.~\ref{tab:performance_ab_grouped}, whereas statistical pairwise model comparisons across different sensor locations (wrist, trunk, and upper arm) using OMC are reported in Tab.~\ref{tab:stats-explo-location}.

\begin{table*}[!htbp]
\centering
\renewcommand{\arraystretch}{1.25}
\caption{Able-bodied classification performance across sensing configurations using wrist (W), trunk (T), and upper-arm (UA) information. Metrics are reported as mean $\pm$ standard deviation.}
\label{tab:performance_ab_grouped}

\resizebox{\textwidth}{!}{%
\begin{tabular}{p{2.2cm} c c ccc ccc ccc}
\toprule
\textbf{Configuration} & $\mathbf{F1_{\text{macro}}}$ & \textbf{MCC}
& \multicolumn{3}{c}{\textbf{Calib}}
& \multicolumn{3}{c}{\textbf{Mov: No TC}}
& \multicolumn{3}{c}{\textbf{Mov: TC}} \\
\cmidrule(lr){4-6} \cmidrule(lr){7-9} \cmidrule(lr){10-12}
& & & Prec & Rec & F1 & Prec & Rec & F1 & Prec & Rec & F1 \\
\midrule

\begin{tabular}[c]{@{}l@{}}IMU\\(W+T)\end{tabular}
& 0.80 $\pm$ 0.07 & 0.73 $\pm$ 0.08
& 0.92 $\pm$ 0.06 & 0.86 $\pm$ 0.13 & 0.88 $\pm$ 0.07
& 0.85 $\pm$ 0.08 & 0.81 $\pm$ 0.12 & 0.82 $\pm$ 0.07
& 0.64 $\pm$ 0.21 & 0.82 $\pm$ 0.14 & 0.68 $\pm$ 0.11 \\

\begin{tabular}[c]{@{}l@{}}OMC\\(W+T+UA)\end{tabular}
& 0.86 $\pm$ 0.05 & 0.81 $\pm$ 0.08
& 0.95 $\pm$ 0.03 & 0.92 $\pm$ 0.08 & 0.93 $\pm$ 0.05
& 0.88 $\pm$ 0.07 & 0.88 $\pm$ 0.07 & 0.88 $\pm$ 0.06
& 0.75 $\pm$ 0.12 & 0.83 $\pm$ 0.13 & 0.78 $\pm$ 0.09 \\

\begin{tabular}[c]{@{}l@{}}OMC\\(W+T)\end{tabular}
& 0.85 $\pm$ 0.06 & 0.79 $\pm$ 0.09
& 0.95 $\pm$ 0.04 & 0.91 $\pm$ 0.10 & 0.92 $\pm$ 0.05
& 0.89 $\pm$ 0.07 & 0.84 $\pm$ 0.12 & 0.86 $\pm$ 0.08
& 0.72 $\pm$ 0.17 & 0.86 $\pm$ 0.10 & 0.77 $\pm$ 0.10 \\

\begin{tabular}[c]{@{}l@{}}OMC\\(T)\end{tabular}
& 0.66 $\pm$ 0.12 & 0.50 $\pm$ 0.13
& 0.72 $\pm$ 0.08 & 0.68 $\pm$ 0.26 & 0.65 $\pm$ 0.16
& 0.73 $\pm$ 0.14 & 0.62 $\pm$ 0.11 & 0.66 $\pm$ 0.10
& 0.60 $\pm$ 0.24 & 0.83 $\pm$ 0.16 & 0.65 $\pm$ 0.17 \\

\begin{tabular}[c]{@{}l@{}}OMC\\(W)\end{tabular}
& 0.78 $\pm$ 0.06 & 0.71 $\pm$ 0.08
& 0.95 $\pm$ 0.03 & 0.91 $\pm$ 0.10 & 0.92 $\pm$ 0.06
& 0.86 $\pm$ 0.07 & 0.77 $\pm$ 0.09 & 0.81 $\pm$ 0.06
& 0.54 $\pm$ 0.13 & 0.77 $\pm$ 0.09 & 0.62 $\pm$ 0.09 \\

\bottomrule
\end{tabular}%
}
\end{table*}

\begin{table*}[!htbp]
\centering
\caption{Pairwise models comparison (using wrist (W), trunk (T), and upper-arm (UA) information) using Wilcoxon signed-rank tests (Holm-corrected p-values). Sig.* corresponds to $p_{\mathrm{corr}} < 0.05$}
\label{tab:stats-explo-location}

\begin{tabular}{ll l c c c}
\toprule
Model A & Model B & Class & $p$ & $p_{\mathrm{corr}}$ & Sig. \\
\midrule
W+T+UA & W+T & Calib        & 0.232 & 1.000 &  \\
W+T+UA & W+T & Mov: No TC   & 0.193 & 1.000 &  \\
W+T+UA & W+T & Mov: TC      & 0.625 & 1.000 &  \\
\\[0.5pt]
W+T+UA & T   & Calib        & 0.002 & 0.035 & * \\
W+T+UA & T   & Mov: No TC   & 0.002 & 0.035 & * \\
W+T+UA & T   & Mov: TC      & 0.004 & 0.039 & * \\
\\[0.5pt]
W+T+UA & W   & Calib        & 0.322 & 1.000 &  \\
W+T+UA & W   & Mov: No TC   & 0.002 & 0.035 & * \\
W+T+UA & W   & Mov: TC      & 0.002 & 0.035 & * \\
\\[0.5pt]
W+T & T & Calib        & 0.002 & 0.035 & * \\
W+T & T & Mov: No TC   & 0.002 & 0.035 & * \\
W+T & T & Mov: TC      & 0.004 & 0.039 & * \\
\\[0.5pt]
W+T & W & Calib        & 0.922 & 1.000 &  \\
W+T & W & Mov: No TC   & 0.084 & 0.588 &  \\
W+T & W & Mov: TC      & 0.004 & 0.039 & * \\
\\[0.5pt]
T & W & Calib        & 0.002 & 0.035 & * \\
T & W   & Mov: No TC   & 0.002 & 0.035 & * \\
T & W   & Mov: TC      & 0.375 & 1.000 &  \\
\bottomrule
\end{tabular}
\end{table*}

\section{Clinical Dataset Model Evaluation}
\label{sec:appendix_performance_patient}
Clinical generalization results using sensors placed on the most affected arm during task execution are presented in Tab.~\ref{tab:patient_affected_results}.
\begin{table*}[!htbp]
\centering
\renewcommand{\arraystretch}{1.25}
\caption{Clinical generalization results using sensor placed on the most affected arm, during tasks involving most affected limb.}

\resizebox{\textwidth}{!}{%
\begin{tabular}{lccc ccc ccc ccc}
\toprule
\textbf{Patient} & $\mathbf{F1_{\text{macro}}}$ & \textbf{MCC} & $\mathbf{AUC_{\text{macro}}}$
& \multicolumn{3}{c}{\textbf{Calib}}
& \multicolumn{3}{c}{\textbf{Mov: No TC}}
& \multicolumn{3}{c}{\textbf{Mov: TC}} \\
\cmidrule(lr){5-7} \cmidrule(lr){8-10} \cmidrule(lr){11-13}
& & &
& Prec & Rec & F1
& Prec & Rec & F1
& Prec & Rec & F1 \\
\midrule

P03 (Right) & 0.51 & 0.35 & 0.85
& 0.15 & 0.34 & 0.21
& 0.90 & 0.96 & 0.93
& 0.67 & 0.27 & 0.38 \\

P18 (Left) & 0.40 & 0.28 & 0.73
& 0.96 & 0.39 & 0.55
& 0.34 & 0.91 & 0.50
& 0.39 & 0.09 & 0.15 \\

P63 (Left) & 0.34 & 0.26 & 0.82
& 0.96 & 0.15 & 0.27
& 0.40 & 0.99 & 0.57
& 0.96 & 0.11 & 0.20 \\

P31 (Right) & 0.45 & 0.25 & 0.72
& 0.86 & 0.31 & 0.44
& 0.48 & 0.85 & 0.61
& 0.28 & 0.34 & 0.30 \\

\midrule
\textbf{Avg $\pm$ Std} & 0.43$\pm$0.06 & 0.28$\pm$0.04 & 0.78$\pm$0.06
& 0.73$\pm$0.35 & 0.30$\pm$0.09 & 0.37$\pm$0.14
& 0.53$\pm$0.22 & 0.93$\pm$0.06 & 0.65$\pm$0.16
& 0.57$\pm$0.28 & 0.20$\pm$0.11 & 0.26$\pm$0.09 \\

\bottomrule
\end{tabular}%
}
\label{tab:patient_affected_results}

\end{table*}

\FloatBarrier   % optional but very helpful
\clearpage      % recommended
\section*{REFERENCES}

\bibliographystyle{IEEEtran}
\bibliography{ieee-bibliography}

\begin{thebibliography}{10}
\providecommand{\url}[1]{#1}
\csname url@samestyle\endcsname
\providecommand{\newblock}{\relax}
\providecommand{\bibinfo}[2]{#2}
\providecommand{\BIBentrySTDinterwordspacing}{\spaceskip=0pt\relax}
\providecommand{\BIBentryALTinterwordstretchfactor}{4}
\providecommand{\BIBentryALTinterwordspacing}{\spaceskip=\fontdimen2\font plus
\BIBentryALTinterwordstretchfactor\fontdimen3\font minus \fontdimen4\font\relax}
\providecommand{\BIBforeignlanguage}[2]{{%
\expandafter\ifx\csname l@#1\endcsname\relax
\typeout{** WARNING: IEEEtran.bst: No hyphenation pattern has been}%
\typeout{** loaded for the language `#1'. Using the pattern for}%
\typeout{** the default language instead.}%
\else
\language=\csname l@#1\endcsname
\fi
#2}}
\providecommand{\BIBdecl}{\relax}
\BIBdecl

\bibitem{oflaherty_2024}
D.~O'Flaherty and K.~Ali, ``Recommendations for upper limb motor recovery: An overview of the {UK} and european rehabilitation after stroke guidelines (2023),'' \emph{Healthcare}, vol.~12, no.~14, p. 1433, 2024.

\bibitem{levin_2009}
M.~Levin, J.~Kleim, and S.~Wolf, ``What do motor ``recovery'' and ``compensation'' mean in patients following stroke?'' \emph{Neurorehabilitation and Neural Repair}, vol.~23, pp. 313--319, May 2009.

\bibitem{jones_2017}
T.~Jones, ``Motor compensation and its effects on neural reorganization after stroke,'' \emph{Nature Reviews Neuroscience}, vol.~18, Mar. 2017.

\bibitem{bokkyu_2024}
B.~Kim, J.~Girnis, V.~Sweet, T.~Nobiling, T.~Agag, and C.~Neville, ``Impact of motor task conditions on end-point kinematics and trunk movements during goal-directed arm reach,'' \emph{Scientific Reports}, vol.~14, Feb. 2024.

\bibitem{murphy_2015}
M.~Alt~Murphy and C.~Hager, ``Kinematic analysis of the upper extremity after stroke---how far have we reached and what have we grasped?'' \emph{Physical Therapy Reviews}, vol.~20, pp. 137--155, May 2015.

\bibitem{brami_2003}
A.~Roby-Brami, A.~Feydy, M.~Combeaud, E.~Biryukova, B.~Bussel, and M.~Levin, ``Motor compensation and recovery for reaching in stroke patients,'' \emph{Acta Neurologica Scandinavica}, vol. 107, pp. 369--381, Jun. 2003.

\bibitem{see_2013}
J.~See, L.~Dodakian, C.~Chou, V.~Chan, A.~McKenzie, D.~Reinkensmeyer, and S.~Cramer, ``A standardized approach to the {Fugl-Meyer} assessment and its implications for clinical trials,'' \emph{Neurorehabilitation and Neural Repair}, vol.~27, Jun. 2013.

\bibitem{wolf_2001}
S.~L. Wolf, P.~A. Catlin, M.~Ellis, A.~L. Archer, B.~Morgan, and A.~Piacentino, ``Assessing {Wolf Motor Function Test} as outcome measure for research in patients after stroke,'' \emph{Stroke}, vol.~32, no.~7, pp. 1635--1639, 2001.

\bibitem{levin_2004}
M.~Levin, J.~Desrosiers, D.~Beauchemin, N.~Bergeron, and A.~Rochette, ``Development and validation of a scale for rating motor compensations used for reaching in patients with hemiparesis: The {Reaching Performance Scale},'' \emph{Physical Therapy}, vol.~84, pp. 8--22, Feb. 2004.

\bibitem{duff_2014}
S.~Duff, J.~He, M.~Nelsen, C.~Lane, V.~Rowe, S.~Wolf, and A.~Dromerick, ``Interrater reliability of the {Wolf Motor Function Test-Functional Ability Scale}: Why it matters,'' \emph{Neurorehabilitation and Neural Repair}, vol.~29, Oct. 2014.

\bibitem{wang_2021}
X.~Wang, Y.~Fu, B.~Ye, J.~Babineau, Y.~Ding, and A.~Mihailidis, ``Technology-based compensation assessment and detection of upper extremity activities of stroke survivors: Systematic review,'' \emph{Journal of Medical Internet Research}, vol.~24, p. e34307, Jun. 2022.

\bibitem{avarell_2022}
E.~Averell, D.~Knox, and F.~van Wijck, ``A real-time algorithm for the detection of compensatory movements during reaching,'' \emph{Journal of Rehabilitation and Assistive Technologies Engineering}, vol.~9, p. 20556683221117085, Sep. 2022.

\bibitem{coias_2022}
A.~C{\'o}ias, M.~Lee, and A.~Bernardino, ``A low-cost virtual coach for {2D} video-based compensation assessment of upper extremity rehabilitation exercises,'' \emph{Journal of NeuroEngineering and Rehabilitation}, vol.~19, Jul. 2022.

\bibitem{zhi_2017}
Y.~Zhi, M.~Lukasik, M.~Li, E.~Dolatabadi, R.~Wang, and B.~Taati, ``Automatic detection of compensation during robotic stroke rehabilitation therapy,'' \emph{IEEE Journal of Translational Engineering in Health and Medicine}, vol.~PP, pp. 1--1, Dec. 2017.

\bibitem{nordin_2016}
N.~B. Nordin, S.~Q. Xie, and B.~C. W{\"u}nsche, ``Simple torso model for upper limb compensatory assessment after stroke,'' in \emph{Proc. IEEE Int. Conf. Advanced Intelligent Mechatronics ({AIM})}, 2016, pp. 775--780.

\bibitem{lin_2023}
H.-P. Lin, L.~Zhao, D.~Woolley, X.~Zhang, H.-J. Cheng, W.~Liang, C.~Kuah, T.~Plunkett, K.~Chua, L.~Zhang, and N.~Wenderoth, ``Exploring the feasibility of computer vision for detecting post-stroke compensatory movements,'' in \emph{Proc. IEEE Int. Conf. Rehabil. Robot. ({ICORR})}, Sep. 2023, pp. 1--6.

\bibitem{unger_2025}
T.~Unger, B.~K{\"u}hnis, L.~Sauerzopf, M.~R. Spiess, A.~de~Spindler, A.~R. Luft, C.~Easthope~Awai, J.~G. Sch{\"o}nhammer, and E.~Gavagnin, ``Using deep learning to detect upper limb compensation in individuals post-stroke using consumer-grade webcams---a feasibility study,'' \emph{Frontiers in Medicine}, vol.~12, 2025.

\bibitem{held_2018}
J.~Held, B.~Klaassen, A.~Eenhoorn, B.-J. van Beijnum, J.~Buurke, P.~Veltink, and A.~Luft, ``Inertial sensor measurements of upper-limb kinematics in stroke patients in clinic and home environment,'' \emph{Frontiers in Bioengineering and Biotechnology}, vol.~6, Apr. 2018.

\bibitem{schwarz_2020}
A.~Schwarz, M.~M.~C. Bhagubai, G.~Wolterink, J.~P.~O. Held, A.~R. Luft, and P.~H. Veltink, ``Assessment of upper limb movement impairments after stroke using wearable inertial sensing,'' \emph{Sensors}, vol.~20, no.~17, p. 4770, 2020.

\bibitem{schwarz_2021}
A.~Schwarz, J.~Veerbeek, J.~Held, J.~Buurke, and A.~Luft, ``Measures of interjoint coordination post-stroke across different upper limb movement tasks,'' \emph{Frontiers in Bioengineering and Biotechnology}, vol.~8, p. 620805, Jan. 2021.

\bibitem{Wittmann2016}
F.~Wittmann, J.~P. Held, O.~Lambercy, M.~L. Starkey, A.~Curt, R.~H{\"o}ver, R.~Gassert, A.~R. Luft, and R.~R. Gonzenbach, ``Self-directed arm therapy at home after stroke with a sensor-based virtual reality training system,'' \emph{Journal of NeuroEngineering and Rehabilitation}, vol.~13, no.~1, Aug. 2016.

\bibitem{nguyen_2021}
G.~Nguyen, J.~MacLean, and L.~Stirling, ``Quantification of compensatory torso motion in post-stroke patients using wearable inertial measurement units,'' \emph{IEEE Sensors Journal}, vol.~PP, pp. 1--1, Apr. 2021.

\bibitem{beer_1999}
R.~F. Beer, J.~D. Given, and J.~P.~A. Dewald, ``Task-dependent weakness at the elbow in patients with hemiparesis,'' \emph{Archives of Physical Medicine and Rehabilitation}, vol.~80, no.~7, pp. 766--772, 1999.

\bibitem{mccrea_2002}
P.~H. McCrea, J.~J. Eng, and A.~J. Hodgson, ``Biomechanics of reaching: Clinical implications for individuals with acquired brain injury,'' \emph{Disability and Rehabilitation}, vol.~24, pp. 534--541, Aug. 2002.

\bibitem{dewald_1995}
J.~P.~A. Dewald, P.~S. Pope, J.~D. Given, T.~S. Buchanan, and W.~Z. Rymer, ``Abnormal muscle coactivation patterns during isometric torque generation at the elbow and shoulder in hemiparetic subjects,'' \emph{Brain}, vol. 118, no.~2, pp. 495--510, 1995.

\bibitem{ellis_2017}
M.~D. Ellis, I.~Schut, and J.~P.~A. Dewald, ``Flexion synergy overshadows flexor spasticity during reaching in chronic moderate to severe hemiparetic stroke,'' \emph{Clinical Neurophysiology}, vol. 128, no.~7, pp. 1308--1314, 2017.

\bibitem{johansson_2015}
G.~M. Johansson, H.~Grip, and C.~Hager, ``Introducing a standardized {Nine Hole Peg Test} in persons with stroke---kinematic analysis,'' \emph{Gait \& Posture}, vol.~42, p. S57, Dec. 2015.

\bibitem{lyle_1982}
R.~C. Lyle, ``A performance test for assessment of upper limb function in physical rehabilitation treatment and research,'' \emph{International Journal of Rehabilitation Research}, vol.~4, no.~4, pp. 483--492, Dec. 1981.

\bibitem{jebsen_1969}
R.~H. Jebsen, N.~Taylor, R.~B. Trieschmann, M.~J. Trotter, and L.~A. Howard, ``An objective and standardized test of hand function,'' \emph{Archives of Physical Medicine and Rehabilitation}, vol.~50, no.~6, pp. 311--319, 1969.

\bibitem{Lhoste2025}
C.~Lhoste, M.~Quast, A.~Ronco, A.~Vogel, C.~E. Awai, M.~Branscheidt, O.~Lambercy, P.~Viskaitis, and D.~Donegan, ``Closed-loop movement-paired transcutaneous auricular vagus nerve stimulation for upper-limb rehabilitation: A feasibility study,'' Sep. 2025, research Square preprint.

\bibitem{barth_2020}
J.~Barth, J.~Klaesner, and C.~Lang, ``Relationships between accelerometry and general compensatory movements of the upper limb after stroke,'' \emph{Journal of NeuroEngineering and Rehabilitation}, vol.~17, p. 138, Oct. 2020.

\bibitem{sauerzopf_2024}
L.~Sauerzopf, C.~G.~C. Panduro, A.~R. Luft, B.~K{\"u}hnis, E.~Gavagnin, T.~Unger, C.~E. Awai, J.~G. Sch{\"o}nhammer, J.~Degenfellner, and M.~R. Spiess, ``Evaluating inter- and intra-rater reliability in assessing upper limb compensatory movements post-stroke: Creating a ground truth through video analysis?'' \emph{Journal of NeuroEngineering and Rehabilitation}, vol.~21, no.~1, Dec. 2024.

\bibitem{ladig_2022}
D.~Laidig and T.~Seel, ``{VQF}: Highly accurate {IMU} orientation estimation with bias estimation and magnetic disturbance rejection,'' \emph{Information Fusion}, vol.~91, pp. 187--204, 2023.

\bibitem{shap_2017}
S.~M. Lundberg and S.-I. Lee, ``A unified approach to interpreting model predictions,'' in \emph{Advances in Neural Information Processing Systems ({NeurIPS})}, 2017, pp. 4768--4777.

\bibitem{chen_2016}
T.~Chen and C.~Guestrin, ``{XGBoost}: A scalable tree boosting system,'' in \emph{Proc. ACM SIGKDD Int. Conf. Knowledge Discovery and Data Mining ({KDD})}, 2016, pp. 785--794.

\bibitem{Chicco2020}
D.~Chicco and G.~Jurman, ``The advantages of the {Matthews} correlation coefficient ({MCC}) over {F1} score and accuracy in binary classification evaluation,'' \emph{BMC Genomics}, vol.~21, no.~1, Jan. 2020.

\bibitem{cirstea_2000}
C.~Cirstea and M.~Levin, ``Compensatory strategies for reaching in stroke,'' \emph{Brain}, vol. 123, no.~5, pp. 940--953, Jun. 2000.

\bibitem{brami_2021}
A.~Roby-Brami, N.~Jarrasse, and R.~Parry, ``Impairment and compensation in dexterous upper-limb function after stroke: From the direct consequences of pyramidal tract lesions to behavioral involvement of both upper-limbs in daily activities,'' \emph{Frontiers in Human Neuroscience}, vol.~15, Jun. 2021.

\bibitem{ranganathan_2017}
R.~Ranganathan, R.~Wang, B.~Dong, and S.~Biswas, ``Identifying compensatory movement patterns in the upper extremity using a wearable sensor system,'' \emph{Physiological Measurement}, vol.~38, no.~12, pp. 2222--2234, Nov. 2017.

\bibitem{ding_2024}
K.~Ding, J.~Wang, X.~Wang, L.~Zhou, D.~Xiong, and L.~Guo, ``System for detection and quantitative evaluation of compensatory movement in post-stroke patients based on wearable sensor and machine learning algorithm,'' \emph{IEEE Sensors Journal}, vol.~24, no.~14, pp. 22\,830--22\,842, 2024.

\bibitem{ranganathan_2017_2}
R.~Ranganathan, R.~Wang, R.~Gebara, and S.~Biswas, ``Detecting compensatory trunk movements in stroke survivors using a wearable system,'' in \emph{Proc. Workshop Wearable Systems and Applications}, Jun. 2017, pp. 29--32.

\bibitem{cai_2020}
S.~Cai, G.~Li, E.~Su, X.~Wei, S.~Huang, K.~Ma, H.~Zheng, and L.~Xie, ``Real-time detection of compensatory patterns in patients with stroke to reduce compensation during robotic rehabilitation therapy,'' \emph{IEEE Journal of Biomedical and Health Informatics}, vol.~24, no.~9, pp. 2630--2638, 2020.

\bibitem{cai_2020_2}
S.~Cai, X.~Wei, E.~Su, W.~Wu, H.~Zheng, and L.~Xie, ``Online compensation detecting for real-time reduction of compensatory motions during reaching: A pilot study with stroke survivors,'' \emph{Journal of NeuroEngineering and Rehabilitation}, vol.~17, Apr. 2020.

\bibitem{seo_2024}
N.~J. Seo, K.~Coupland, C.~Finetto, and G.~Scronce, ``Wearable sensor to monitor quality of upper limb task practice for stroke survivors at home,'' \emph{Sensors}, vol.~24, p. 554, Jan. 2024.

\bibitem{berjis_2025}
M.~Berjis, M.-E. LeBel, D.~Lizotte, and A.~Trejos, ``Selecting muscles for detection of upper-limb compensatory movements using s-{EMG} sensors,'' \emph{IEEE Transactions on Medical Robotics and Bionics}, vol.~PP, pp. 1--1, Jan. 2025.

\end{thebibliography}

\end{document}